\documentclass[letterpaper, journal]{ieeetran}

\IEEEoverridecommandlockouts                              

\usepackage{amsmath,amsfonts}
\usepackage{algorithmic}
\usepackage{array}
\usepackage[caption=false,font=normalsize,labelfont=sf,textfont=sf]{subfig}
\usepackage{textcomp}
\usepackage{stfloats}
\usepackage{url}
\usepackage{graphicx}
\usepackage{hyperref}
\usepackage[T1]{fontenc}
\def\BibTeX{{\rm B\kern-.05em{\sc i\kern-.025em b}\kern-.08em
    T\kern-.1667em\lower.7ex\hbox{E}\kern-.125emX}}
\usepackage{balance}

\begin{document}


\title{Learning Multi-Agent Task Assignment and Navigation in the Factory: from Simulation to Real Robots
}

\author{Abdalwhab Bakheet Mohamed Abdalwhab$^{1}$, Giovanni Beltrame$^{2}$, David St-Onge$^{1}$ 
\thanks{
This project was funded by NSERC Alliance grant ALLRP 566182 - 21 and NSERC CREATE CoRoM program.
}
\thanks{$^{1}$ INIT Robots Lab, École de technologie supérieure, 1100 Notre-Dame W., Canada {\tt\footnotesize abdalwhab-bakheet-mohamed.abdalwhab.1@ens.etsmtl.ca}}
\thanks{$^{2}$ MISTLab, Department of Computer and Software Engineering, Polytechnique Montréal, Canada}
}

\maketitle
\thispagestyle{empty} 
\pagestyle{empty} 

\begin{abstract}

Reinforcement learning (RL) has shown considerable promise for robotic
decision-making, yet deploying multi-agent RL (MARL) on physical multi-robot
systems in industrial environments remains challenging. This paper investigates
the real-world applicability of decentralized MARL for multi-robot multi-machine
tending. We propose Feature-fusion Multi-Agent Proximal Policy Optimization
(FMAPPO), which fuses 2D LiDAR measurements with task-specific state information
to enable safe decentralized multi-robot task assignment and navigation. A complete simulation-to-reality pipeline was developed using high-fidelity robotic simulation and ROS2 and deployed on physical mobile-manipulator platforms operating under realistic real-world conditions, with the robotic arms disabled during the experiments. We further investigate the sensitivity of the learned policy to command update frequency, an important consideration for real-world deployment. Comparative evaluation in simulation demonstrated that FMAPPO significantly outperformed state-of-the-art baselines with a large effect size, achieving improvements of 106\% and 21\% in parts delivery and 48\% and 11\% in parts collection over MAPPO and SMAPPO, respectively. FMAPPO also increased machine utilization by 31 and 10 percentage points, respectively, while reducing collisions by 18\% and 15\% and increasing the safety score by 14 and 6 percentage points compared with MAPPO and SMAPPO, respectively. Furthermore, real-world experiments demonstrated that the learned decentralized policies can coordinate multiple robots to service multiple machines while maintaining safe operation under real-world sensing and control constraints. Videos of the real-world experiment are available online\footnote{\url{https://anonymouspapers123.github.io/FMAPPO/}}.

\end{abstract}


\section{Introduction}

Recent advances in Multi-agent reinforcement learning (MARL) have demonstrated impressive capabilities in coordination tasks, most notably in benchmark environments such as StarCraft II \cite{vinyals2019grandmaster}, Google Research Football \cite{kurach2020google}, and the OpenAI Multi-Agent Particle Environment (MPE) \cite{lowe2017multi}.

Combining MARL with robotics has the potential to help reduce the manpower shortage in the manufacturing sector by enabling more flexible automation. While recent studies have demonstrated the potential of MARL for many robotic domains such as exploration, navigation, transportation, and task allocation \cite{orr2023multi}, the reliable deployment of those solutions remains a major challenge. In fact, very limited multi-robot navigation research has reached the step of validation in real robots, and mainly in lab settings \cite{tang2025deep}.

Machine tending is an excellent example of such a repetitive, labor-intensive task that autonomous robots can handle. A fleet of decentralized mobile manipulators can autonomously tend to a group of production machines, freeing human workers to focus on more cognitively demanding tasks. Robots can navigate between machines, feeding raw material and taking ready parts from machines to the designated storage areas. However, managing multiple robots that tend to multiple machines can be challenging because it requires coordination, task assignment, navigation, and collision avoidance on top of temporal reasoning \cite{AbdalwhabSclable2026}.  

Few previous researchers have explored machine tending using reinforcement learning, and often focus on single-robot~\cite{iriondo2019pick,iriondo2023learning} or rely on highly simplified simulation setups that do not capture the complexity of real robotic systems~\cite{agrawal2021multi,abdalwhab2025attention,AbdalwhabSclable2026}. More work is needed to reduce the gap between simplified simulations and real-world complexity and to investigate the deployability of MARL solutions on real robotic platforms.  

To contribute to filling this gap, we present a decentralized MARL framework for multi-robot multi-machine tending and evaluate its performance in both simulation and physical deployment. To the best of our knowledge, this is the first work to investigate decentralized MARL for machine tending in a high-fidelity simulator using realistic mobile manipulator models handling both task allocation and navigation. It is also, to the best of our knowledge, the first to demonstrate successful deployment of that setup on physical robots using onboard sensing while operating under realistic constraints. In summary, the main contributions of this study are:

\begin{itemize}
    \item Framework contribution: A sensor-grounded decentralized MARL formulation enabling joint task assignment and continuous navigation to transfer zero-shot from simulation to physical multi-robot systems.

    \item Methodological contribution: Feature-fusion representation combining task semantics with local LiDAR geometry.

    \item Physical robot-learning contribution: Demonstration and quantitative characterization of zero-shot transfer under real sensing, localization, communication, and actuation constraints.

    \item Deployment insight: The learned policy is sensitive to command update frequency, with simulation and real-world performance exhibiting different frequency-dependent trends, highlighting the importance of evaluating update frequencies for real-world deployment.

\end{itemize}

\section{Related Work}\label{sec:RelatedWorks}

High-fidelity simulators such as Isaac Sim~\cite{gao2026nvidia} increasingly provide accurate representations of robot dynamics and environments, helping reduce the simulation-to-reality (sim-to-real) gap. Nevertheless, discrepancies in sensing and control, such as differences in observation and action delays, can remain between simulation and physical deployment. Several studies have therefore incorporated such effects into policy learning~\cite{neto2026reinforcement}. For example, Bouteiller et al.~\cite{bouteiller2021reinforcement} augmented observations with sensing and action delays and action history, and enhanced Soft Actor-Critic with a multi-step value estimator and partial trajectory resampling. Their approach was evaluated using a delay-augmented version of the Gym MuJoCo continuous control suite. Sandha et al.~\cite{sandha2021sim2real} instead augmented observations with execution time and sampling intervals and trained policies under varying delays, demonstrating sim-to-real transfer on a physical DeepRacer car. In contrast, our work does not explicitly model or compensate for these differences during training. Instead, we consider them as part of the sim-to-real gap and evaluate the robustness of the learned policy to them. The following subsections review related work on reinforcement learning with LiDAR observations for navigation and reinforcement learning for machine tending.

\subsection{RL with LiDAR Data for Navigation}

Previous work has incorporated LiDAR measurements, including both 2D
scans~\cite{lai2025dare,de2024spatiotemporal,nagar2024reinforcement,abdalmanan20232d}
and 3D point
clouds~\cite{Zhang2025,chen2025virtual,xu2025flying,li2025multi,liu2026temporally},
into reinforcement learning (RL) frameworks for navigation. Existing approaches
generally follow one of two architectural paradigms. The first employs a
separate perception module that processes the LiDAR data into a compact
intermediate representation before passing it to the RL
policy~\cite{Zhang2025,li2025multi}. The second directly feeds the LiDAR
observations, either alone~\cite{chen2025virtual,lai2025dare} or fused with
additional sensor modalities and state
information~\cite{xu2025flying,liu2026temporally}, into the policy network for
action prediction.

Zhang et al.~\cite{Zhang2025} proposed a graph-based perception framework for autonomous exploration that combines 3D LiDAR observations with the Soft Actor-Critic (SAC) algorithm. Their method converts raw point clouds into a sparse, informative graph representation using graph attention networks, and feeds it to the RL policy to predict exploration waypoints, while conventional path planning and low-level controllers generate the corresponding robot motions. Similarly, Li et al.~\cite{li2025multi} designed a dedicated perception pipeline that fuses camera, LiDAR, and IMU measurements to estimate road boundaries, semantic segments, and surrounding object poses. These high-level features are processed by a cross-domain attention module before being provided to a Deep Deterministic Policy Gradient (DDPG)-based policy for autonomous driving.

On the other hand, other works perform end-to-end policy learning directly from LiDAR observations. Chen et al.~\cite{chen2025virtual} proposed a DDPG-based autonomous driving framework in which 3D LiDAR point clouds are processed using a PointNet architecture enhanced with self-attention, allowing the actor network to directly predict steering and acceleration commands. For aerial navigation, Xu et al.~\cite{xu2025flying} employed PPO using 3D LiDAR together with IMU measurements and a local state estimator. Their framework extracts latent features from the LiDAR data and combines them with the estimated vehicle state before predicting the desired thrust and body rate. Liu et al.~\cite{liu2026temporally} also adopted a PPO-based architecture for UAV navigation, where convolutional neural networks encode the 3D LiDAR observations before fusing them with the UAV state and handcrafted geometric features derived from both the LiDAR measurements and the goal position, such as obstacle clearance and goal-direction visibility. To further improve safety, they apply a shielding mechanism that suppresses motion commands directed toward nearby obstacles.

LiDAR has also been successfully applied in RL for collision avoidance using only 2D range measurements. Lai et al.~\cite{lai2025dare} proposed DARE, an enhanced SAC framework that learns a latent representation of the LiDAR observations through supervised learning auxiliary objectives, including collision-risk prediction, next-state prediction, and reward estimation. The learned latent representation is combined with the raw LiDAR scan and waypoints generated by a global planner before being passed to the policy network, targeting navigation in cluttered environments.

In general, previous LiDAR-based RL works have shown that incorporating LiDAR measurements can enhance navigation performance and collision avoidance. However, they mainly focus on single-robot setups where the policy needs to reason only about navigation and obstacle avoidance. In contrast, our model targets decentralized multi-robot task assignment and navigation, where the policy has to reason about task assignment, navigation, collision avoidance, and interaction with other robots, demanding consolidated task-specific representation learning.

\subsection{RL for Machine Tending}\label{p3subsec:RlforMT}

Despite the success of reinforcement learning in many robotic applications \cite{singh2022reinforcement}, its investigation for machine tending with mobile robots remains relatively limited. Iriondo et al.~\cite{iriondo2019pick,iriondo2023learning} learn optimal positioning of the base of a mobile manipulator using feedback from the manipulator planner. Their approach evolved from DDPG and PPO to using TD3, reporting good results on a physical robot. However, the study was limited to a single part retrieval without considering part availability state, transporting the part to the drop-off locations.

The increased complexity of multi-robot manufacturing has led most existing works to address only subsets of the overall problem or to rely on simplified assumptions. Li et al.~\cite{li2025self} focused on raw material delivery and proposed a three-network architecture: one centralized task assignment policy, a waypoint selection policy, and a value estimation network. However, they relied on a simplified grid-based simulation environment for evaluation and assumed predetermined task execution times. In addition, their centralized task assignment limits the system scalability. Similarly, Ho et al.~\cite{ho2025integrated} formulated task assignment as a centralized DQN problem, also assuming fixed task durations.

More holistic approaches have attempted to jointly optimize task allocation and navigation. The pioneering work of Agrawal et al.~\cite{agrawal2021multi} developed a PPO-based framework that considers machine processing delays and failures during job scheduling and robot navigation. Nevertheless, their approach assumes unlimited robot carrying capacity and relies on a centralized communication architecture with marginal centralized control. Siddiqua et al.~\cite{siddiqua2024information} integrated PPO with an information-sharing mechanism that encodes the locations of robots, packages, and waste into separate occupancy grids. However, their formulation also assumes infinite onboard storage and does not evaluate collision avoidance or safety performance.

Abdalwhab et al.~\cite{abdalwhab2025attention} introduced an attention-based observation encoder integrated with MAPPO to jointly learn task allocation and navigation for multi-machine tending without the assumption of infinite onboard storage. This architecture was subsequently extended to support varying numbers of robots, machines, and storage locations without requiring network modifications or retraining~\cite{AbdalwhabSclable2026}. Although these works reduced the assumptions and demonstrated improved scalability, they were validated only in simplified simulation environments.

In contrast, this work bridges the gap between multi-agent RL simulation and real-world deployment. To the best of our knowledge, it is the first study to investigate decentralized MARL for machine tending in a high-fidelity simulator using realistic mobile manipulator models, handling joint task allocation and navigation, while also demonstrating successful deployment and experimental validation on physical robots.

\section{Problem Formulation}\label{sec:ProblemFormulation} 

Similar to \cite{abdalwhab2025attention}, we consider the problem of multi-agent, multi-machine tending, where a team of $N$ autonomous mobile manipulators services a manufacturing facility containing $M$ production machines and $S$ storage locations. The objective of the robots is to maximize manufacturing throughput by collecting finished parts from machines and transporting them to the appropriate storage areas while ensuring safe operation. 

As machines complete processing asynchronously, robots must continuously make decentralized decisions regarding task selection, navigation, and execution. During operation, robots are required to avoid collisions with other robots, machines, and storage areas while efficiently reaching their assigned destinations. After a finished part is collected, the corresponding machine
immediately starts processing the next workpiece, requiring a fixed processing time $T$ before another part becomes available. In this work, we consider the case of two robots, two machines, and one storage area ($N=2, M=2, S=1$), and focus on assessing the deployability of the solution.

Reliable object pose estimation, grasp planning, and object placement are fundamental components of a complete machine-tending system. In this work, however, our primary objective is decentralized multi-robot coordination, task allocation, navigation, and event sequencing, with an emphasis on achieving robust and safe operation on physical robotic platforms. To isolate these challenges from manipulation-specific issues, we adopt the same assumptions as in~\cite{abdalwhab2025attention,AbdalwhabSclable2026}, namely that robots perform pick-and-place operations continuously without stopping and that the replenishment of raw materials is performed by an external process. This allows the proposed framework to concentrate exclusively on the coordination, navigation, and scheduling challenges that dominate large-scale autonomous manufacturing systems.

\section{Methods}\label{p3sec:Methodology}

\subsection{Feature Fusion MAPPO (FMAPPO) }\label{subsec:FMAPPO}

We use MAPPO~\cite{yu2022surprising} as our backbone. MAPPO is a multi-agent reinforcement learning algorithm that works in a paired actor-critic setup. The actor learns to take actions to maximize the reward, while the critic estimates the advantage of those actions to guide the actor's learning. Following the centralized training with decentralized execution (CTDE) paradigm, the critic has access to the global state and is only used during training, whereas the policy is executed using local observations alone.

\begin{figure}
    \centering
    \includegraphics[width=0.9\linewidth]{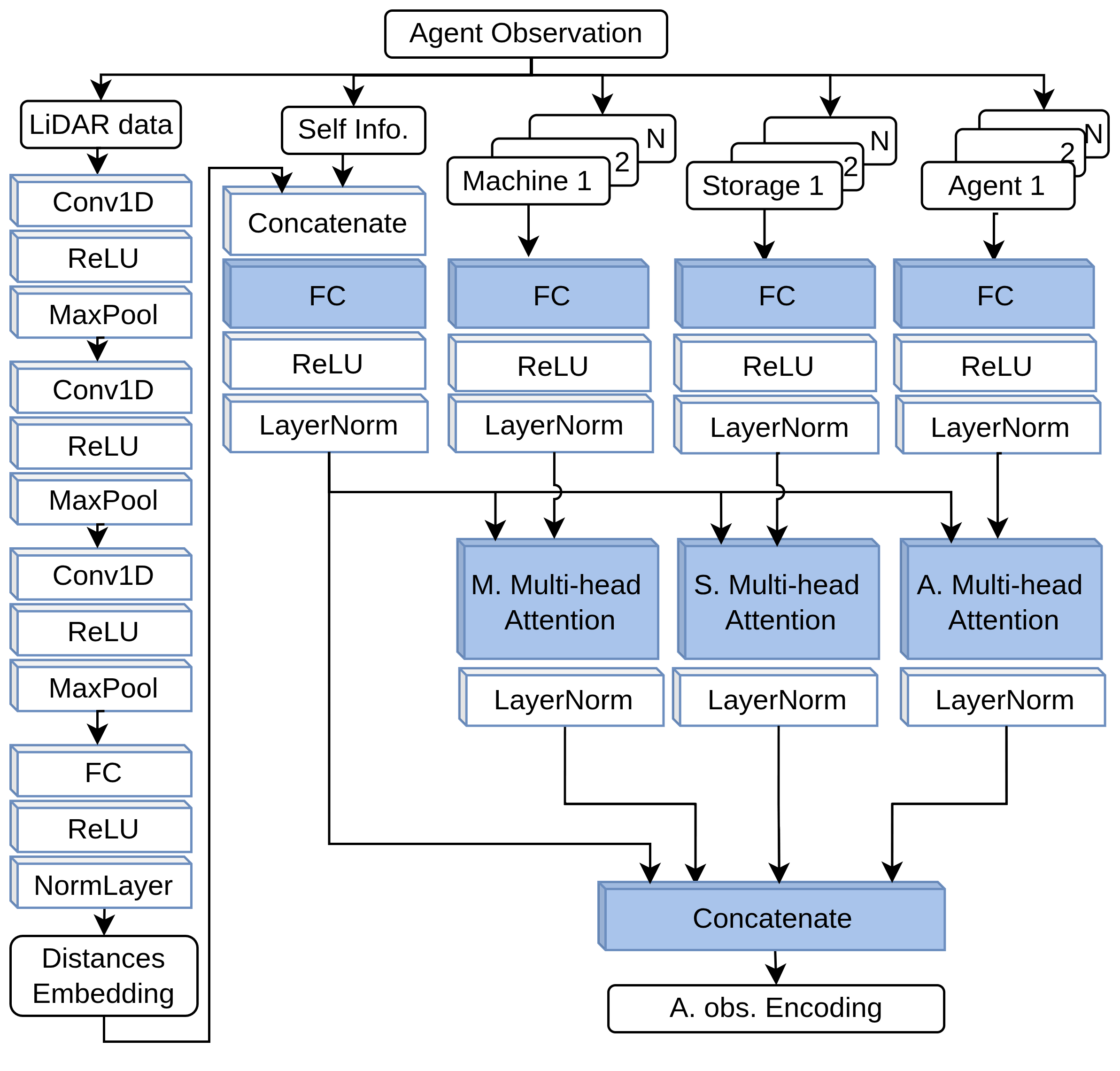}
    \caption{Our Proposed Feature-Fusion Observation Encoder with the LiDAR processing branch.}
    \label{fig:Observation_Encoder}
\end{figure}

The policy must preserve the semantic structure of the task state while extracting local geometric information from substantially higher-dimensional LiDAR observations. Directly treating these inputs as a homogeneous observation vector can obscure their different roles in decision-making. We therefore introduce a feature-fusion encoder in which geometric and task-semantic observations are processed independently before being combined. Building on the entity-based observation encoder of~\cite{AbdalwhabSclable2026}, we introduce our observation encoder illustrated in Figure~\ref{fig:Observation_Encoder}.

The downsampled LiDAR scan is represented as a vector of distance measurements normalized between zero and one and processed independently from the remaining observation components. The LiDAR branch consists of three blocks of one-dimensional convolutional layers with ReLU activations and max-pooling operations to progressively extract local spatial features from the scan. The resulting feature map is then passed through a fully connected layer followed by a ReLU activation and layer normalization to produce a compact LiDAR embedding.

Different semantic components of the observation are encoded independently to allow the network to learn specialized feature representations for each entity type. Each entity encoder consists of a fully connected layer followed by a ReLU activation and layer normalization. The LiDAR embedding is concatenated with the agent's position and velocity before being processed by the agent encoder to generate the agent embedding. Likewise, the machines, storage areas, and neighboring robots' observations are encoded independently in parallel.

To aggregate information from multiple entities, a separate multi-head attention module is applied to each entity category. The agent embedding serves as the query, while the corresponding entity embeddings act as keys and values. This allows the network to selectively attend to the entities that are most relevant to the current robot state, producing a single context vector for the machines, storage areas, and neighboring robots, respectively. Finally, the agent embedding and the three context vectors are concatenated to form the final observation embedding.

The actor network uses the observation encoder architecture to produce the observation embedding, then processes it using three fully connected blocks, each consisting of a linear layer, ReLU activation, and layer normalization. A final linear layer predicts the mean of the continuous action distribution, while the logarithm of the action standard deviation is maintained as a learnable parameter. During training, actions are sampled from the resulting Gaussian distribution to encourage exploration. During evaluation, the deterministic mean action is used directly.

Unlike the actor, the centralized critic has access to the observations of all agents. Therefore, the observation encoder is applied independently to every agent in parallel to obtain a set of agent observation embeddings. A multi-head attention layer with a learnable query aggregates these embeddings into a fixed-length representation of the global state, followed by layer normalization. This state representation is then processed by three fully connected blocks with ReLU activations before a final linear layer estimates the state value.

\subsection{Observation, Reward, and Action}\label{subsec:ObsRandA}

\textbf{Our observation} design was inspired by the work \cite{AbdalwhabSclable2026, abdalwhab2025attention}. Similar to them, each robot observation included the following:
\begin{itemize}
    \item The robot's global position, with a binary indicator specifying whether it is carrying a part.
    \item The relative locations of all machines, accompanied by their current part-availability status.
    \item The relative position of the designated drop-off area.
    \item The relative positions of the remaining robots in the fleet, along with binary indicators denoting whether each one is carrying a part.
\end{itemize}

In addition to that, to enhance the model's avoidance capability, we added the following to the observation:
\begin{itemize}
    \item The robot's z-axis orientation and linear velocities.
    \item The processed distance points from the robot's two LiDARs.
    \item The relative orientation and linear velocities of the other robots.
\end{itemize}

LiDAR points are processed by downsampling, resetting points beyond the LiDAR maximum detection distance to the LiDAR maximum detection distance, and then normalized to the range [0,1].  

\textbf{Reward Function:} Our reward formulation is based on the reward structure proposed in \cite{AbdalwhabSclable2026}. We retain and tuned the part collection reward ($R_{pi}$), part delivery reward ($R_{pl}$), distance-based shaping rewards, namely progress toward a machine ($R_{pm}$) and progress toward the drop-off area ($R_{ps}$), as well as the time penalty term ($R_t$).

To improve collision-avoidance learning, we modified the collision reward design introduced in \cite{AbdalwhabSclable2026}. Rather than assigning the same penalty throughout a collision event, we distinguish between initiating a collision and remaining in contact. The motivation is that, for real-world deployment, preventing collisions is more critical than merely recovering from them after they occur. Consequently, the collision penalty is decomposed into two components: a new-collision penalty ($R_{nc}$), applied when a collision first occurs, and a collision-duration penalty ($R_{dc}$), applied at every timestep while the collision persists. The penalties are designed such that $R_{nc} > R_{dc}$, thereby strongly discouraging collision initiation.

Furthermore, we refined the uncollected-parts penalty proposed in \cite{AbdalwhabSclable2026}. Instead of penalizing all robots equally, the penalty is applied only to robots that are not currently carrying a part. This modification encourages idle robots to service waiting machines while avoiding unnecessary penalties for robots already engaged in transportation tasks:

\begin{equation}\label{equ:R_u}
    R_{u} = \begin{cases}
                s_u * p_{un} / M & \text{if the robot has no part}\\
                0 & \text{otherwise}
            \end{cases}
\end{equation}

where $s_u$ is a scaling coefficient, $p_{un}$ denotes the total number of uncollected parts, and $M$ is the number of machines.

The overall reward is obtained by summing all reward components:

\begin{equation}\label{equ:RT3}
R_T = R_{pi} + R_{pl} + R_{nc} + R_{dc} + R_{pm} + R_{ps} + R_u + R_t.
\end{equation}

\textbf{Actions}: The action space consists of continuous linear velocity commands along the robot's (x)- and (y)-axes. Although continuous control increases the complexity of the learning problem compared to discrete action spaces used in \cite{AbdalwhabSclable2026}, it more accurately reflects the kinematic capabilities of a holonomic mobile robot and enables smoother, more natural navigation behaviors.

Although the action space excludes angular velocity commands, the robot's orientation is included in the observation space. In both simulation and real-world experiments, we noticed that slight heading drift can occur during motion, and collisions in simulation may also alter the robot's orientation. Including the heading enables the policy to account for these changes. Moreover, many mobile manipulation platforms require a specific robot orientation for successful manipulation, making this information beneficial for generalization.  

\subsection{Simulation}\label{subsec:Sim}
NVIDIA Isaac Sim \cite{gao2026nvidia} was selected as the simulation platform due to its high-fidelity, GPU-accelerated physics engine, while Isaac Lab \cite{mittal2025isaaclab} was used for its comprehensive support for multi-agent reinforcement learning research and development. Building upon these frameworks, we developed a novel multi-robot, multi-machine tending reinforcement learning environment in which a fleet of N mobile manipulators navigates among M machines to collect completed parts and transport them to a designated drop-off area.

Similar to \cite{AbdalwhabSclable2026}, the primary focus of this work is the integration of task allocation and navigation rather than manipulation. Therefore, manipulation actions are abstracted within the simulation: when a robot reaches a machine containing a completed part, the part is assumed to be successfully collected, and when the robot reaches the drop-off area, the part is assumed to be successfully delivered.

\begin{figure}[!t]
    \centering
    \includegraphics[width=0.8\linewidth]{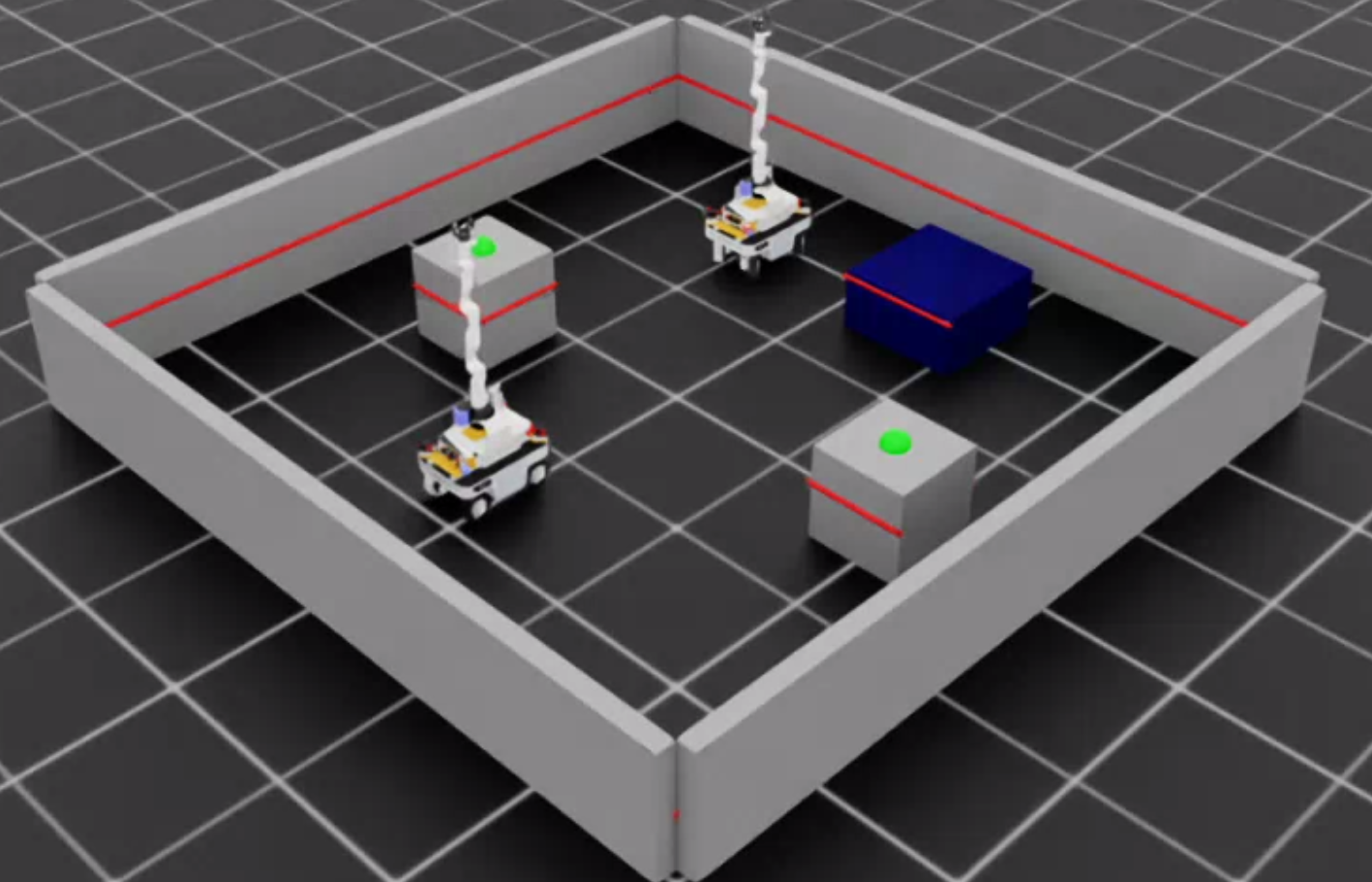}
    \caption{Example of a randomly generated layout in Isaac Sim. The environment contains two mobile manipulators (Rangen robots), a drop-off area (blue), and two machines (gray) with green indicators denoting available parts. The red points represent returns from the simulated safety LiDAR sensors mounted on the robots.}
    \label{fig:SimSetup}
\end{figure}

Figure~\ref{fig:SimSetup} illustrates the simulation environment in one of the randomly generated layouts. The figure shows two Rangen robots, each consisting of an AgileX Ranger Mini holonomic mobile base equipped with a Kinova Gen3 manipulator (the arm was not used in this work). The drop-off area is highlighted in blue, while the machines are represented by gray boxes. A green indicator light above a machine denotes the availability of a completed part for collection. The red points correspond to returns from two simulated 270-degree ray-casting sensors, which emulate the safety LiDARs mounted on the physical robots.

The workspace consists of a 6×6-meter arena. At the beginning of each episode, the robots, machines, and drop-off area are randomly initialized. To ensure sufficient spatial diversity while preventing excessively clustered configurations, all entities are placed such that the minimum distance between any pair of locations is 2.1 m. Furthermore, all sampled positions are constrained to lie within a 4.2×4.2 m region centered in the arena, ensuring both adequate separation and consistent interaction opportunities throughout training.

\subsection{Real Deployment Platform}\label{subsec:Real_Plat}

\begin{figure}
    \centering
    \includegraphics[width=.45\linewidth]{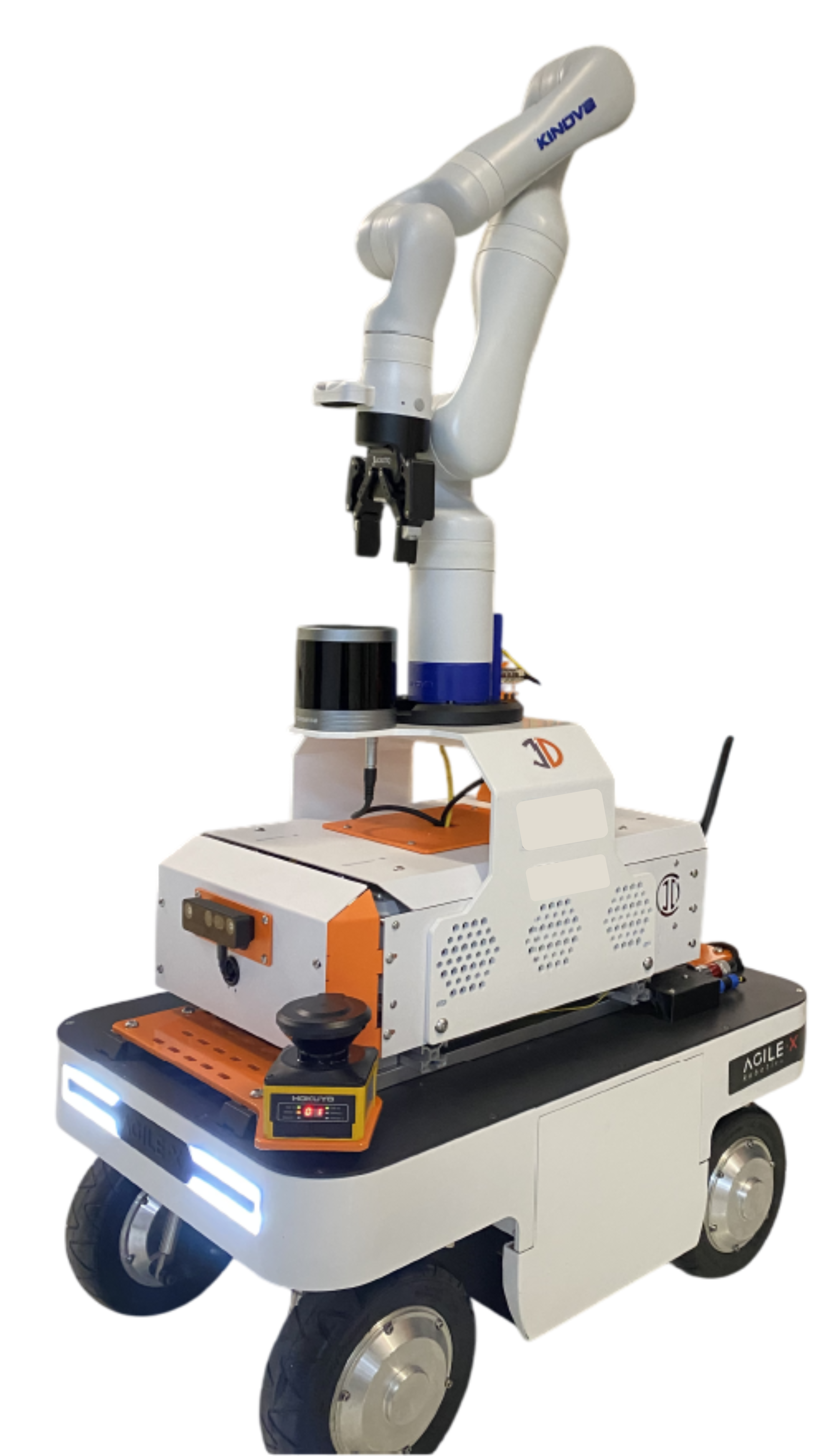}
    \caption{Our Mobile Manipulator (Rangen): AgileX Ranger Mini omnidirectional base equipped with a Kinova Gen 3 arm; this work doesn't include arm control.}
    \label{fig:Rangen}
\end{figure}

The real-world experiments were conducted using two Rangen mobile manipulation platforms (Figure~\ref{fig:Rangen}). Each platform comprises an AgileX Ranger Mini omnidirectional mobile base equipped with a Kinova Gen3 robotic arm (the arm was not used in this work). To enable reliable omnidirectional obstacle perception, two Hokuyo UAM-05LP 2D safety LiDARs are mounted diagonally, providing full 360° environmental coverage. Onboard computation is provided by an NVIDIA Jetson AGX and an Intel NUC, while an integrated network router connects all sensors, the Jetson, and the NUC to the local Wi-Fi network. In addition, the platform is equipped with a Phidgets IMU, a RoboSense Helios 16P 3D LiDAR, and an OAK-D Pro Wide RGB-D camera, supporting a wide range of perception and localization capabilities. 
 
\subsection{Real Deployment Architecture}\label{subsec:Real}

\begin{figure}[!t]
    \centering
    \includegraphics[width=.9\linewidth]{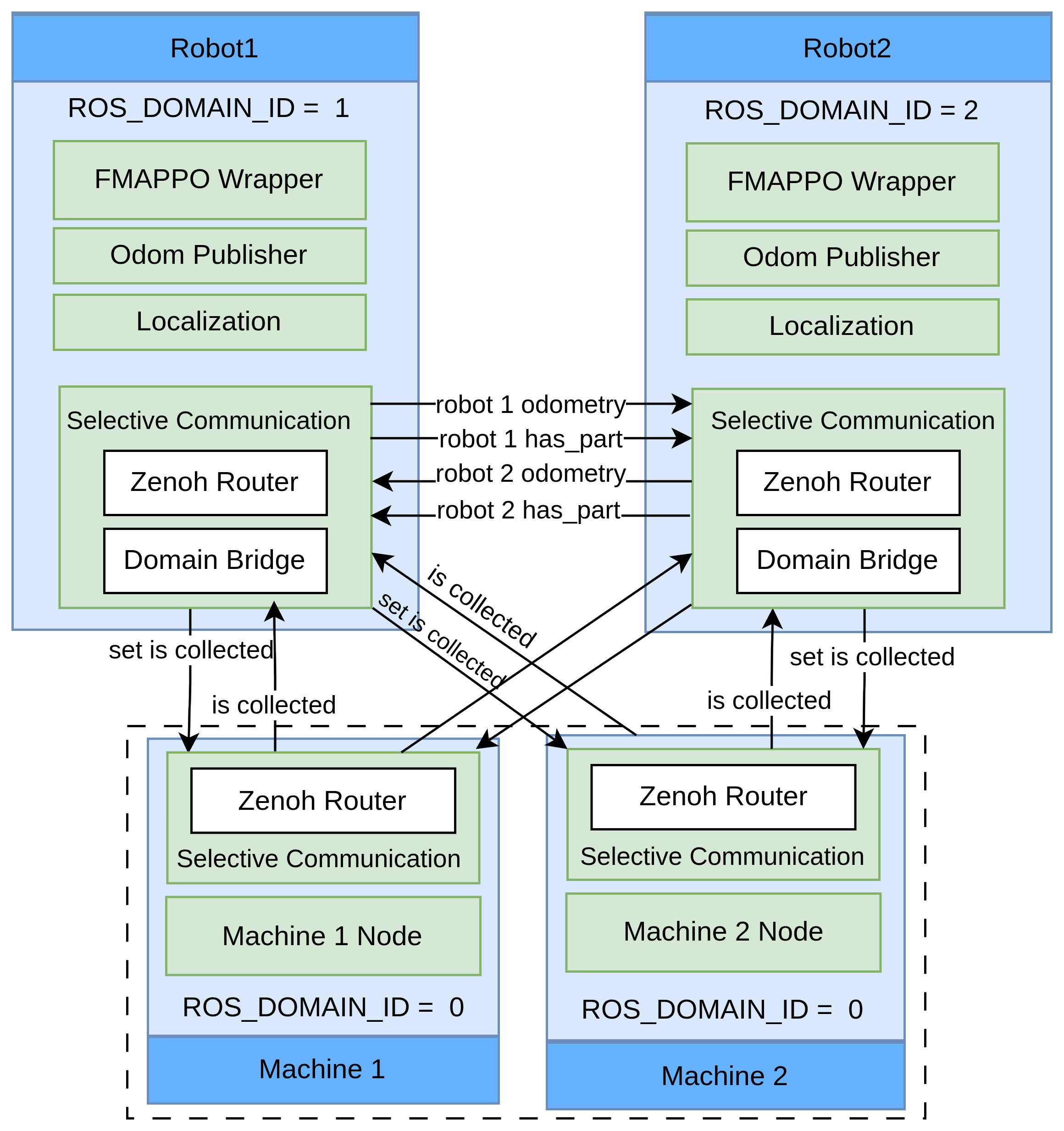}
    \caption{Real deployment setup.}
    \label{fig:deploymentSetup}
\end{figure}

Figure~\ref{fig:deploymentSetup} illustrates the architecture used for the real-world deployment. The system is designed to be fully decentralized; each robot independently performs localization, communication, and policy inference without relying on a central coordinator.

For localization, each robot runs an Extended Kalman Filter (EKF) that fuses wheel odometry with Inertial Measurement Unit (IMU) measurements to estimate the robot pose in the odometry frame and continuously update the ROS TF tree. To further improve localization accuracy, the scans from the two 2D LiDARs are merged and processed by SLAM Toolbox~\cite{Macenski2021}, which performs simultaneous localization and mapping (SLAM) to estimate the robot pose in the world frame and update the TF tree accordingly. The Odom Publisher node then retrieves the latest robot pose in the world frame from the TF tree and combines it with the velocity estimates from the EKF, transformed into the world frame, to produce the final odometry message used by the control policy.

To isolate the software stack running on each robot, every robot operates within its own ROS Domain ID. Only the information required by the policy is exchanged between robots and machines, minimizing communication overhead and improving network scalability. Zenoh~\cite{CorsaroZenoh2023} is used as the communication middleware both within each robot and for communication among robots and machines due to its low communication overhead and flexible publish-subscribe architecture. Each robot and machine hosts its own Zenoh Router, forming a distributed communication network. A ROS Domain Bridge on each robot selectively imports only the required topics from the remote domains, namely the other robot's odometry and \texttt{has\_part} status, together with the \texttt{is\_collected} and \texttt{set\_collected} topics exchanged with each machine.

The FMAPPO Wrapper encapsulates the trained FMAPPO policy and serves as the interface between the ROS ecosystem and the reinforcement learning model. It is responsible for loading the trained policy, collecting observations from the subscribed ROS topics, constructing the policy observation vector, executing model inference, and publishing the predicted velocity commands to the robot low-level controller.

All machines operate within a common ROS Domain ID. In the presented experiments, the two machines were simulated on a single ground-station laptop, while the mobile manipulators consisted of the physical Rangen platforms described in Section~\ref{subsec:Real_Plat}.

\subsection{Evaluation Metrics}\label{subsec:EvalMetrx}

For our simulation, we built on the evaluation metrics introduced in \cite{AbdalwhabSclable2026}:

\begin{itemize}

\item \textbf{Task Throughput}: Measured by the total number of parts collected and total parts delivered by all robots.

\item \textbf{Safety}: Measured by the total number of collision events across all robots. An overlap between a robot's collision circle and another entity's circle or the environment boundary is counted as one collision until the overlap ends. Each robot--robot circle overlap is counted as two collisions (one for each robot); details of the circles' radii are in the additional materials.   

\item \textbf{Machine Utilization}: Defined as the ratio of collected parts to the maximum achievable production:

\begin{equation}\label{equ:MU}
MU = \frac {P_c} {P_{max}}
\end{equation}

\end{itemize}

where $P_c$ represents the total number of parts collected from all machines, and $P_{max}$ is the theoretical maximum number of parts that could be produced by all machines under uninterrupted operation, assuming no delays caused by waiting for robots to collect completed parts.

However, the total number of collisions alone does not fully reflect safety, as an agent may achieve fewer collisions simply by collecting or delivering fewer parts. To account for this trade-off, we additionally define the \textbf{Safety Score} ($ST$), which normalizes collisions by task completion:

        \begin{equation}\label{equ}
             ST = \frac {1} {1 + \frac{C_T}{P_c + P_d}}
        \end{equation}

where $C_T$ is the total number of collisions, and $P_c$ and $P_d$  are the total number of parts collected and the total number of parts delivered. A higher Safety Score indicates safer operation while maintaining higher task throughput.

For the actual deployment, we still monitored the total parts collected and delivered, as in the simulation. In addition, we introduced two more measures, specifically for the real deployment: 

\begin{enumerate}
    \item \textbf{Inter-Robot Separation (d):} measured as the minimum and mean distance between the robots, because the number of collisions can not be used as the comparison metric since the physical robots should not collide. 
    \item \textbf{Navigation quality:} Expressed by path length ($S$), action smoothness ($J_a$) and trajectory smoothness ($J_p$). ($J_a$) is measured by the jerk based on the second derivative of actions predicted by the policy, while ($J_a$) is measured by the jerk computed from the third derivative of the robots' position updates during the navigation trip.
    

\end{enumerate}

\section{Experiments}\label{subsec:Exps}

This section covers the main experiments, including performance comparison, a study on the sensitivity of the learned policy to command update frequency, an ablation study and real-world validation. Furthermore, we refer readers to the additional materials for more results analysis and additional experiments.

\subsection{Performance Comparison}\label{subsec:PerfComp}
To evaluate the performance of the proposed FMAPPO against the baselines MAPPO and SMAPPO, all models were trained under identical conditions for 500K environment interaction steps using 256 parallel environments. Each episode consists of 300 environment steps, with a fixed processing latency of 150 steps introduced for part handling, allowing the two machines to produce a maximum of four parts per episode. Each environment step corresponds to six physics simulation steps, during which the same predicted commands are repeated. 


For evaluation, performance metrics were computed over the final 200 episodes of training, which are treated as the convergence phase of each run. To ensure statistical robustness and reproducibility, all experiments were repeated across seven independent runs, each initialized with a distinct random seed.

Training was conducted on a high-performance workstation equipped with an NVIDIA GeForce RTX 5090 GPU (32 GB VRAM), 24 CPU cores, and 2×32 GB of system memory. Under this configuration, simultaneous execution of up to seven training runs required approximately 3.5 days to complete.

Table~\ref{tab:performance} depicts the performance of the proposed FMAPPO vs MAPPO and SMAPPO, reporting the mean and standard deviation over the seven seeds.

\begin{table}[ht]
\centering
\caption{Performance comparison of FMAPPO vs MAPPO and SMAPPO, when trained for 500K environment steps using 256 parallel environments, with performance metrics computed over the final 200 episodes, reporting the mean (std) over seven different seeds.}
\label{tab:performance}
\begin{tabular}{|c|c|c|c|}
\hline
Metric & MAPPO & SMAPPO & FMAPPO \\ \hline
Collected & 2.50 (0.15) & 3.33 (0.21) & \textbf{3.71 (0.05)} \\ \hline 
Delivered & 1.62 (0.21) & 2.77 (0.34) & \textbf{3.34 (0.07)} \\ \hline 
Collisions & 2.06 (0.09) & 1.99 (0.13) & \textbf{1.69 (0.15)} \\ \hline 
Safety Score & 0.67 (0.02) & 0.75 (0.01) & \textbf{0.81 (0.02)} \\ \hline 
Utilization & 0.62 (0.04) & 0.83 (0.05) & \textbf{0.93 (0.01)} \\ \hline 
\end{tabular}
\end{table}

FMAPPO consistently outperforms both MAPPO and SMAPPO across all evaluated metrics, achieving substantial gains in task efficiency and safety. Specifically, FMAPPO improves parts delivery by 106\% compared with MAPPO and by 21\% compared with SMAPPO. Similarly, it improves parts collection by 48\% and 11\%, respectively. FMAPPO also increases utilization by 31 and 10 percentage points compared with MAPPO and SMAPPO, respectively. From a safety perspective, it reduces the number of collisions by 18\% and 15\%, respectively, while increasing the safety score by 14 and 6 percentage points, respectively. As mentioned previously, each overlap between a robot's collision circle and another entity is counted as a collision, even when no physical contact occurs. Robot-robot circle overlaps are counted as two collisions, one for each robot. Therefore, the reported collision counts provide a conservative estimate and are expected to overestimate the number of actual physical collisions. In addition, since layouts are generated randomly, some configurations may present more challenging navigation conditions than those expected in the real-world layout. To ensure the robustness of these findings, we first applied the Shapiro–Wilk test to verify the normality of the paired differences between the two methods across all metrics. This was followed by a paired-sample t-test, with Holm–Bonferroni correction for multiple comparisons (with a significance level of $\alpha$=0.05) to assess statistical significance. Then, Cohen’s standardized mean difference was computed to quantify effect sizes. Across all metrics, the results indicate statistically significant improvements with large effect sizes, confirming the superiority of FMAPPO over both MAPPO and SMAPPO. Due to space limitations, we included the details of the statistical analysis in the additional materials. 

\begin{figure*}[!t]
    \centering
    \includegraphics[width=.78\linewidth]{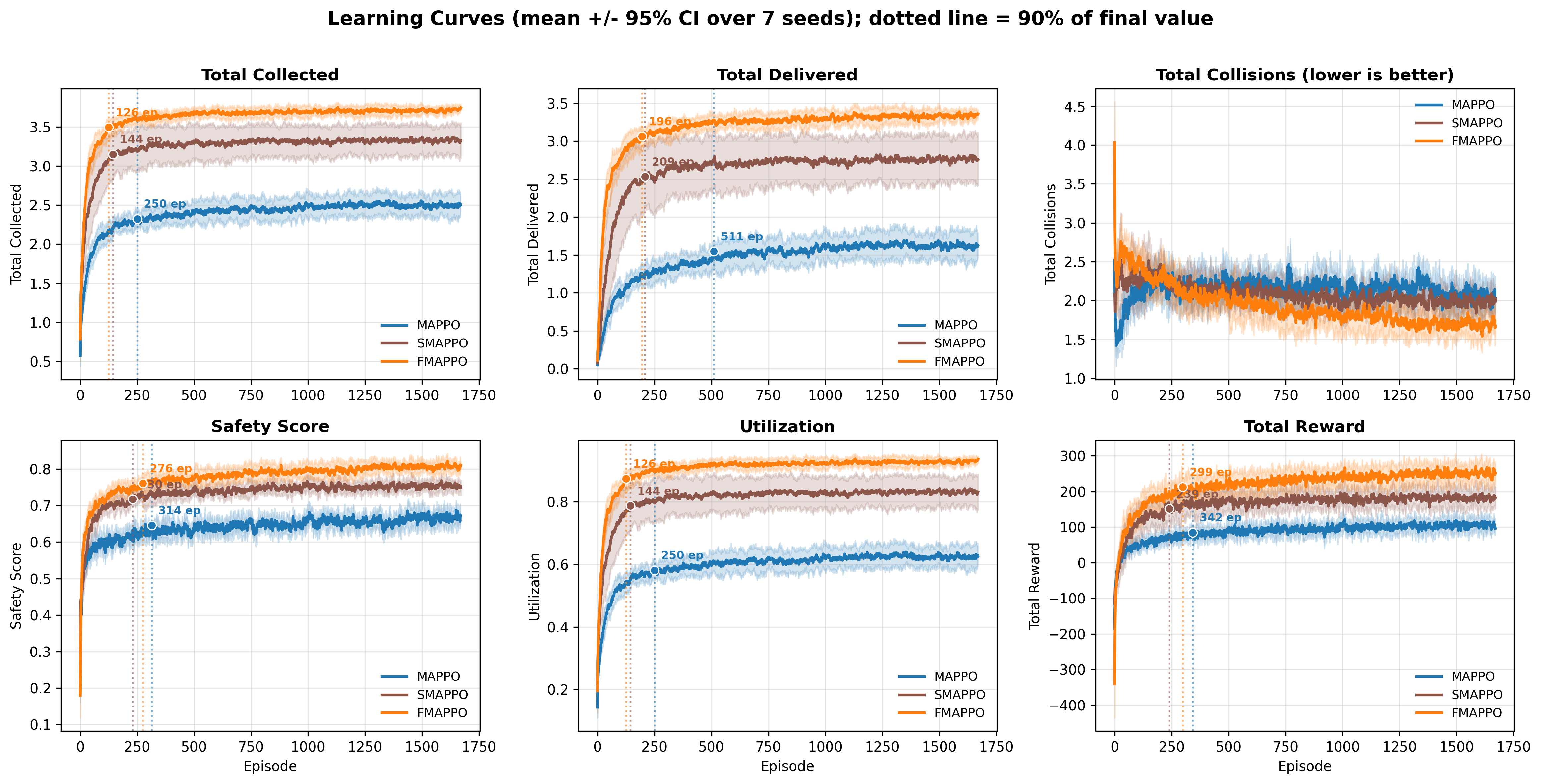}
    \caption{Learning curves for FMAPPO vs MAPPO and SMAPPO}
    \label{fig:FMAPPO_learning}
\end{figure*}

Figure~\ref{fig:FMAPPO_learning} illustrates the learning dynamics of FMAPPO compared to MAPPO and SMAPPO. The curves report the mean performance over seven random seeds (solid lines), with shaded regions indicating the 95\% confidence intervals. From the earliest training episodes, FMAPPO demonstrates superior performance in parts collection, parts delivery, machine utilization, and total reward, with the performance gap steadily widening over time. Although FMAPPO initially exhibits a higher collision rate---likely due to more aggressive exploration and increased task engagement---it rapidly improves in safety performance and surpasses both MAPPO and SMAPPO in collision avoidance at around episode number 250, after which the margin continues to grow.

Moreover, FMAPPO exhibits markedly better sample efficiency and faster learning across all task components. For instance, in parts delivery, it surpasses both MAPPO and SMAPPO's final converged performance in fewer than 100 episodes and reaches 90\% of its own final performance within 196 episodes, demonstrating significantly improved sample efficiency and convergence speed. More about sample efficiency is in the additional materials.

\subsection{Sensitivity to Command Update Frequency}

This experiment provides a comprehensive evaluation of the sensitivity of the learned policy to command update frequency, which is impractical to perform extensively on the physical robots due to the time required for real-world experiments. The same two scenarios of the real deployment were reproduced in Isaac Lab to evaluate the trained policy robustness to command update frequency change and sensor noises. To closely match the real deployment conditions, the simulation incorporated the same system constraints, including a LiDAR update rate of 34.2 Hz and a maximum linear velocity of 0.5 m/s, which was imposed for safety in the real robot's experiments.

To further reduce the sim-to-real gap, realistic sensor and state-estimation noise were incorporated into the simulation. Zero-mean Gaussian noise with a standard deviation of $\sigma=10$ cm was applied to the robot position estimates, reflecting the typical localization accuracy reported for systems such as the Loco Positioning System~\cite{bitcraze_loco_positioning}. Likewise, zero-mean Gaussian noise was added to the estimated linear velocities with $\sigma=10$ cm/s. To account for perception uncertainty, Gaussian noise with $\sigma=1$ cm was injected into the LiDAR range measurements, consistent with the measurement accuracy specified in the Hokuyo UAM-05LP manual~\cite{hokuyo_uam05lp}. Finally, Gaussian perturbations with $\sigma=5$ cm were applied to the robots' initial positions to emulate small errors in the starting position.

\begin{figure}
    \centering
    \includegraphics[width=0.9\linewidth]{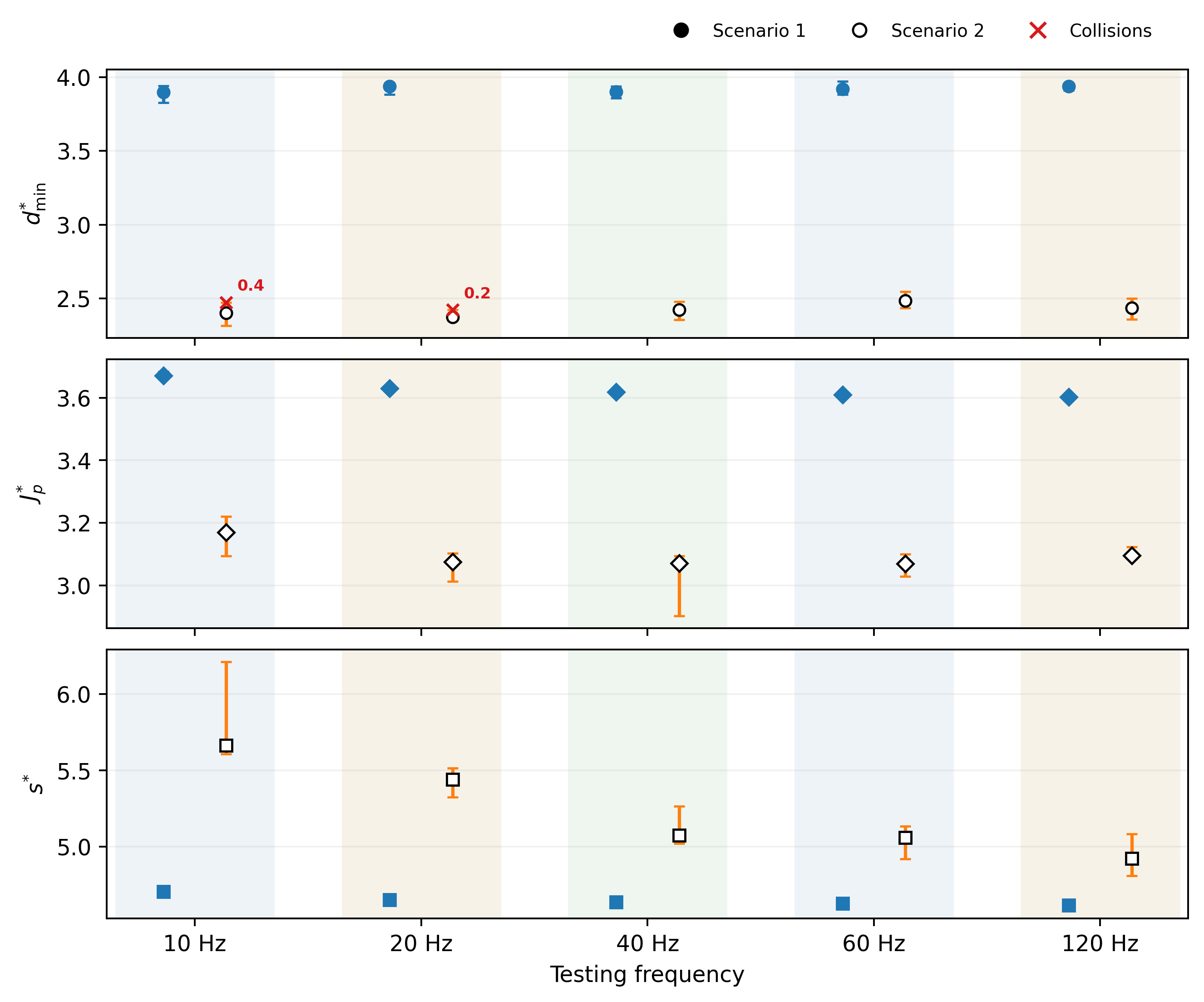}
    \caption{Frequency sensitivity analysis of the simulation study showing dimensionless metrics (Minimum inner robot separation $d^*$, trajectory smoothness $J_p^*$, and path length $S^*$) for each scenario and tested frequency as medians with interquartile ranges, in addition to the average number of collisions (if any)}
    \label{fig:new_sim_freq_sensitivity}
\end{figure}

For each scenario, the same policy that was trained with a command update frequency of 20 Hz was re-evaluated at update frequencies of 10, 20, 40, 60, and 120 Hz, with sensor noise and velocity guard. Each experiment was repeated 10 times using different random seeds. To compare the tested policy frequencies, we followed the same general procedure as the real-world validation experiment, treating frequency as a categorical factor, and analyzing the balanced $2\times5$ subset comprising both scenarios at 10, 20, 40, 60, and 120~Hz ($n=100$). Metrics were also made dimensionless using the same methodology. Figure~\ref{fig:new_sim_freq_sensitivity} shows the dimensionless metrics for each scenario and tested frequency as medians with interquartile ranges, in addition to the average number of collisions for the cases that had any.

Similarly, a two-factor Freedman-Lane residual-permutation ANOVA tested frequency, scenario, and their interaction. Holm correction was applied separately to each effect family across all metrics. Frequency significantly affected mean inter-robot distance ($F(4,90)=3.56$, $p_{\rm perm}=0.0016$, $p_{\rm Holm}=0.0048$, $\eta_p^2=0.137$) and navigation path length ($F(4,90)=2.84$, $p_{\rm perm}<0.001$, $p_{\rm Holm}<0.001$, $\eta_p^2=0.112$), but not minimum inter-robot distance ($F(4,90)=1.35$, $p_{\rm Holm}=0.52$) nor the trajectory smoothness ($F(4,90)=0.08$, $p_{\rm Holm}=1$). Scenario affected minimum inner-robot distance ($F(1,90)=824.4$, $p_{\rm Holm}<0.001$), path length ($F(1,90)=14.24$, $p_{\rm Holm}<0.001$), and trajectory smoothness ($F(1,90)=259.7$, $p_{\rm Holm}<0.001$), while the frequency$\times$scenario interaction affected path length ($F(4,90)=2.396$, $p_{\rm Holm}<0.001$) and mean inner-robot distance ($F(4,90)=3.1226$, $p_{\rm Holm}<0.0102$). We further refer readers to the additional materials for more details, including performance metrics at the whole task level.

\begin{figure}
    \centering
    \includegraphics[width=.9\linewidth]{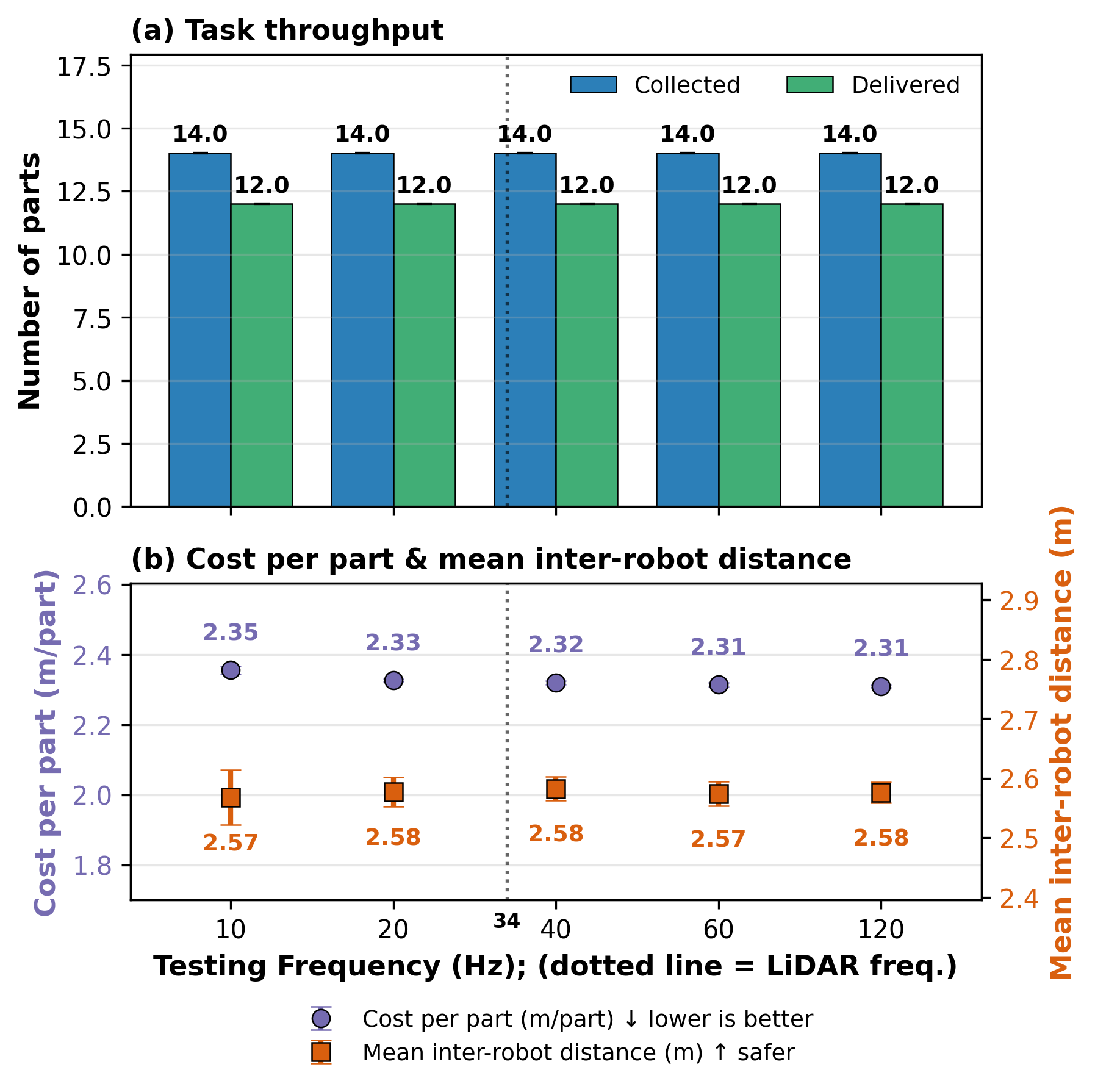}
    \caption{Sensitivity to command update frequency in simulation: overall task  performance metrics: Scenario 1 results, reporting average and std.}
    \label{fig:sim_freq_s1}
\end{figure}

\begin{figure}
    \centering
    \includegraphics[width=.9\linewidth]{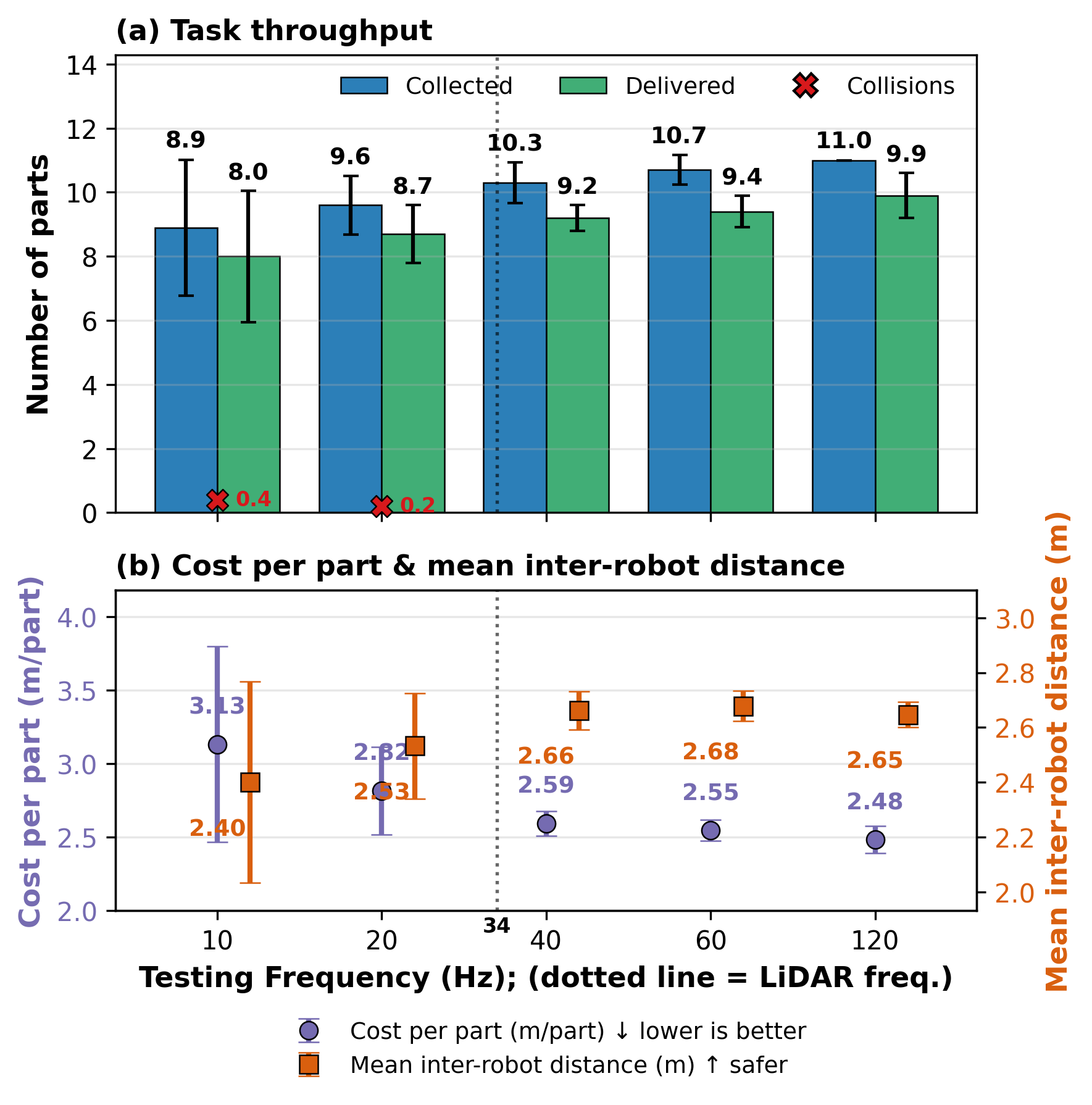}
    \caption{Sensitivity to command update frequency in simulation: overall task  performance metrics: Scenario 2 results, reporting average and std}
    \label{fig:sim_freq_s2}
\end{figure}

Secondly, we report the whole-task level performance metrics. Fig.~\ref{fig:sim_freq_s1} and Fig.~\ref{fig:sim_freq_s2} present the average task throughput, task completion cost ($Cost = TravelDistance_{total} / (parts_{collected}+ parts_{delivered})$), mean inter-robot distance, and number of collisions (if any) for Scenarios 1 and 2, respectively.

In Scenario 1, which is considerably less complex than Scenario 2, varying the command update frequency did not affect the throughput. However, a consistent reduction in the task completion cost was observed as the command frequency increased. Similarly, the mean distance between the two robots generally increased with higher command frequencies, indicating safer behavior, except in the 60~Hz case, which showed a slight decrease. No collisions occurred at any of the evaluated frequencies, indicating that the task remained sufficiently simple for the learned policy to maintain safe operation despite the added noise and changes in the control rate.

In contrast, Scenario 2 exhibited a much stronger dependence on the command update frequency. Higher command frequencies resulted in increased throughput accompanied by lower performance variance across evaluation runs. Moreover, the mean inter-robot distance generally increased while the completion cost consistently decreased as the command frequency was raised. Since the policy was trained in a noise-free simulation, introducing realistic sensor and state-estimation noise occasionally led to collisions at lower command frequencies. As the command frequency approached the training frequency, the number of collisions decreased substantially, and when operated at frequencies higher than those used during training, the policy successfully avoided collisions in all evaluation runs.

We also performed a statistical analysis to evaluate the significance and effect size of the performance differences between command update frequencies of 10 Hz and 120 Hz, with p-value correction for multiple comparisons in each scenario, details of the analysis results are shown in the additional materials. In Scenario 1, the task completion cost exhibited statistically significant differences with large effect sizes. In Scenario 2, the number of collected and delivered parts, as well as the task completion cost and mean inner-robot distance, showed statistically significant improvements with large effect sizes. The remaining metrics did not exhibit statistically significant differences under the evaluated experimental conditions.


\subsection{Ablation Study}\label{subsec:ablation}
This experiment evaluates the design choices made in FMAPPO. To assess the contribution of each component, we compare FMAPPO with four variants: FMAPPO without LiDAR information (NoLID), FMAPPO without velocity information (NoVel), and FMAPPO without orientation information (NoOrient). Table \ref{tab:ablation} presents the results of this ablation study.

\begin{table}[ht]
\centering
\caption{Ablation Study: Final Performance Comparison (FMAPPO, NoLID, NoVel, NoOrient) mean (std)}
\label{tab:ablation}
\begin{tabular}{|c|c|c|c|c|}
\hline
Metric & FMAPPO & NoLID & NoVel & NoOrient \\ \hline
Collected & 3.71 (0.05) & 3.60 (0.07) & 3.70 (0.06) & 3.73 (0.07) \\ \hline 
Delivered & 3.34 (0.07) & 3.17 (0.10) & 3.38 (0.08) & 3.37 (0.11) \\ \hline 
Collisions & \textbf{1.69 (0.15)} & 2.04 (0.15) & 1.91 (0.18) & 1.75 (0.21) \\ \hline 
Safety Score & \textbf{0.81 (0.02)} & 0.77 (0.01) & 0.79 (0.02) & 0.80 (0.02) \\ \hline 
Utilization & \textbf{0.93 (0.01)} & 0.90 (0.02) & 0.92 (0.01) & 0.93 (0.02) \\ \hline 
\end{tabular}
\end{table}


The results indicate that including LiDAR information had a positive impact, enhancing the model's performance in all measured metrics. Furthermore, both LiDAR and velocity information have a considerable impact on collision avoidance, as removing them leads to an increase in the number of collisions by 21\% and 13\%, respectively. In contrast, incorporating the robots' orientation information provides only a marginal benefit in this setup.


\subsection{Real-World Validation}\label{subsec:RealworldVal}

For real-world validation, we deployed the same FMAPPO model trained in simulation directly on two real robots in two different factory-floor layouts. We also wanted to validate the model's robustness and sensitivity to changes in command update frequency. Since, in the physical system, policy inference and command publication are not synchronized with all sensing processes. In particular, the safety LiDARs provide measurements at 34.2 Hz, whereas the learned policy was trained with an action-update rate of 20 Hz and is evaluated here between 20 and 120 Hz. Consequently, at evaluation rates above the LiDAR acquisition frequency, multiple policy command updates may occur between successive range observations. This experiment therefore assesses not only sensitivity to the policy command update rate, but also robustness to the difference in frequencies between sensing and action execution.

To evaluate different policy update frequencies while keeping the robot command interface unchanged, the base command rate was fixed at 120~Hz, and each policy output was repeated for multiple consecutive command cycles before a new action was computed. For both scenarios, the arena measured $6\times6~m$. The robots were initialized at fixed starting positions $r_1$ and $r_2$ and operated for 7200 command cycles, corresponding to one minute, during which they attempted to collect and deliver as many parts as possible.

\begin{figure}[!t]
    \centering
    \includegraphics[width=.9\linewidth]{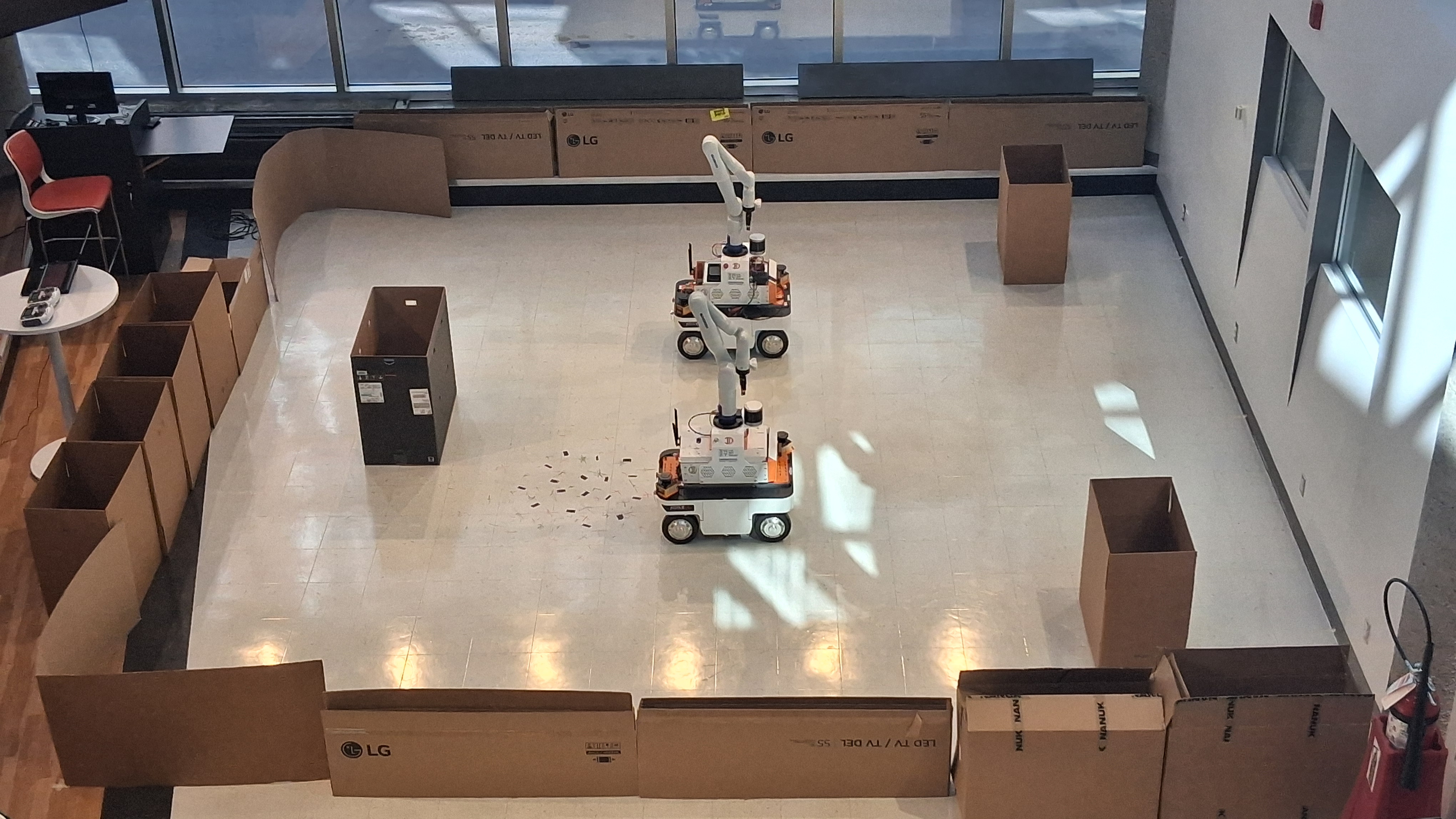}
    \caption{Real deployment first layout: the two brown boxes on the right represent the machines, and the black box in the back represents the drop area.}
    \label{fig:Scenario1_layout}
\end{figure}

Figure~\ref{fig:Scenario1_layout} depicts the first layout with the two robots initialized in the middle of the room, two meters apart from each other. In this layout, while the model was trained with a command update frequency of 20 Hz, we tested once at each of four command update frequencies (20,40,60, and 120 Hz), allowing the two robots to do as many collection and delivery trips as they can in the  7200 timesteps. 

\begin{figure}[!t]
    \centering
    \includegraphics[width=.9\linewidth]{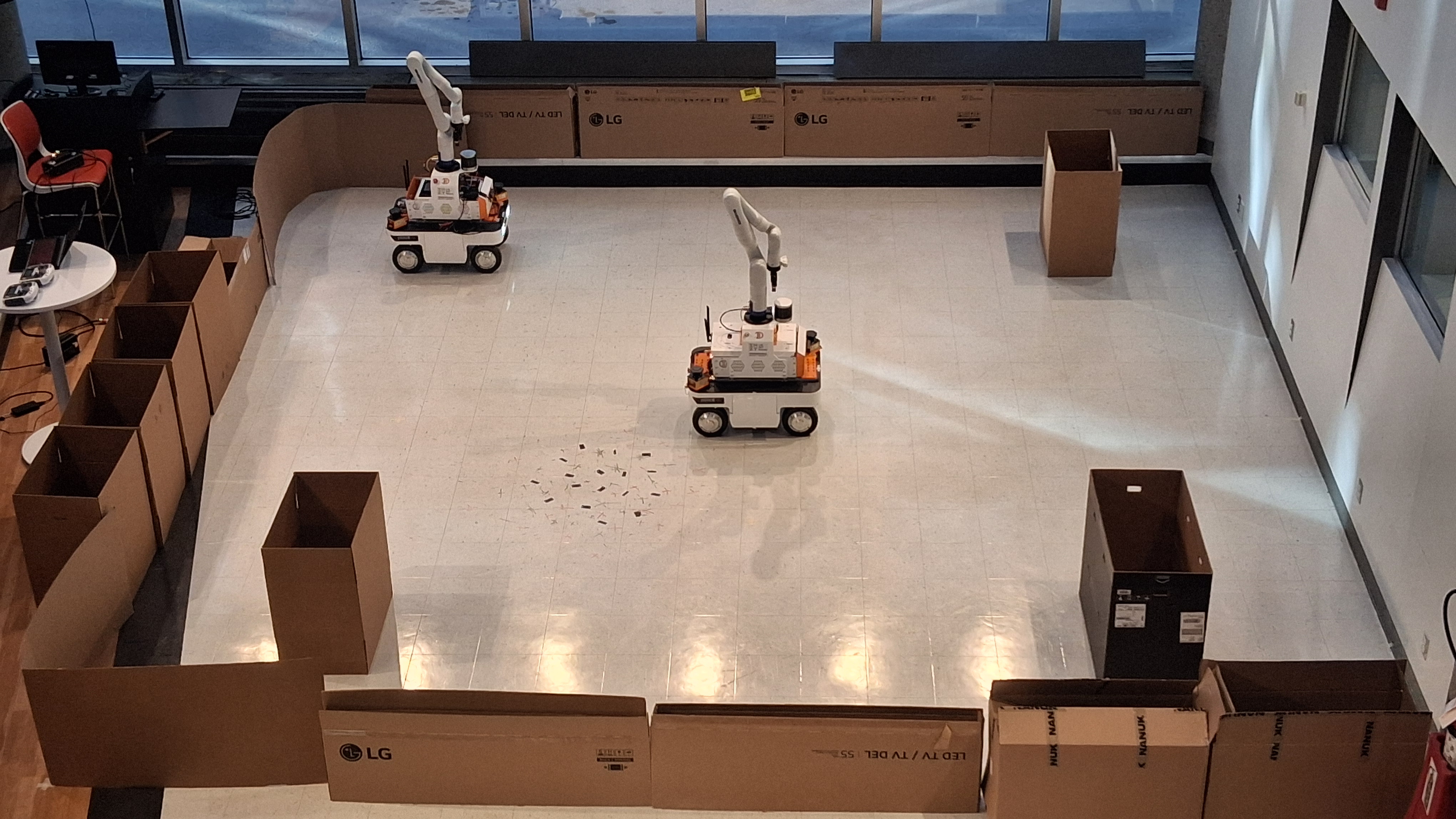}
    \caption{Real deployment second scenario: the two brown boxes on a diagonal represent the machines, and the black box in the bottom right corner represents the drop area.}
    \label{fig:Scenario2_layout}
\end{figure}

For the second layout (figure~\ref{fig:Scenario2_layout}), robots were placed to have the same distance from the two machines, so that it is not obvious which machine each robot should go to. Following the same evaluation procedures used in the first layout, we reevaluated the three best-performing frequencies (20, 40, and 60 Hz) one time each.

\begin{figure}
    \centering
    \includegraphics[width=.9\linewidth]{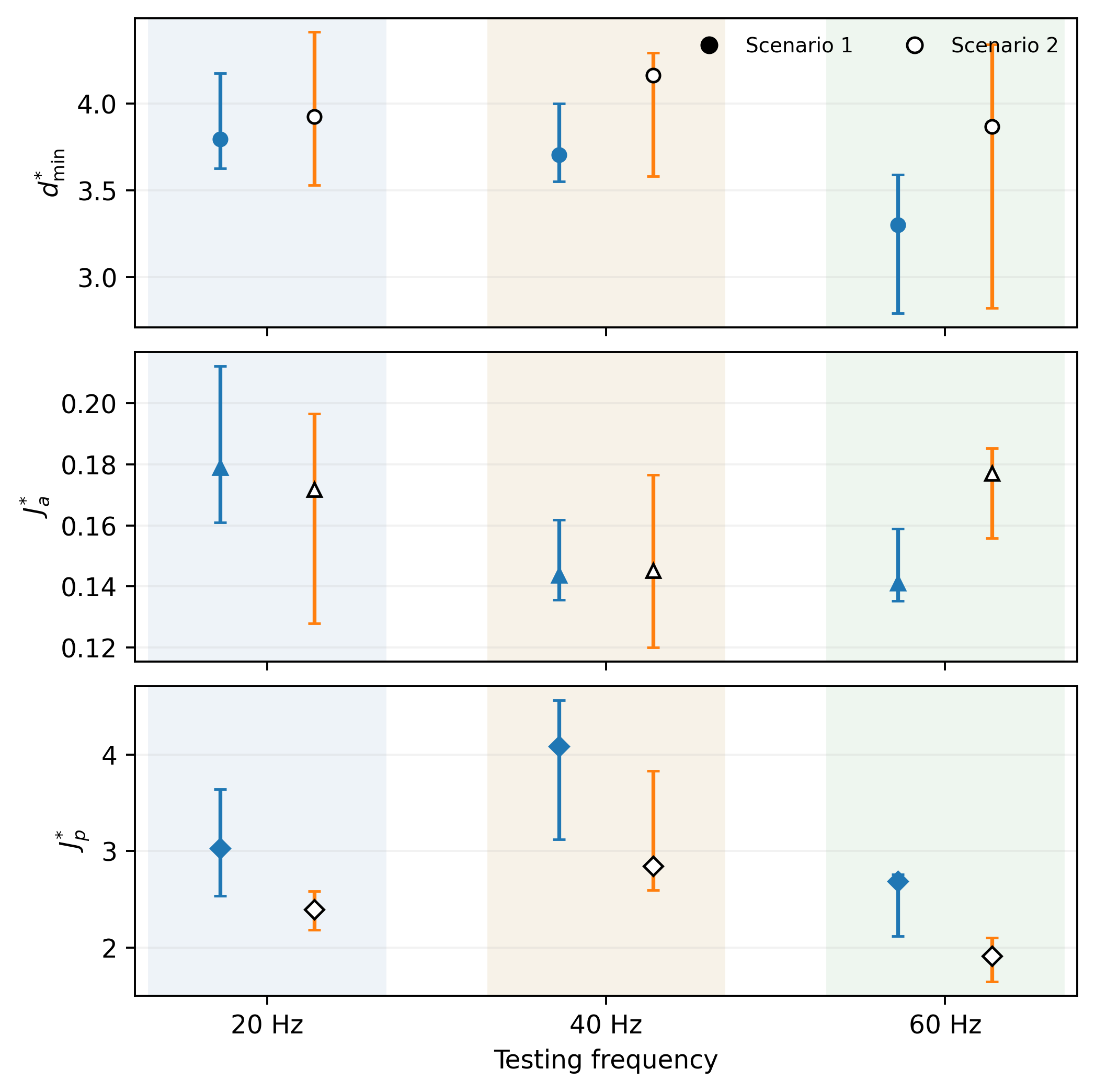}
    \caption{Frequency sensitivity analysis of the real deployment experiment showing dimensionless metrics (Minimum inner robot separation $d^*$, action smoothness $J_a^*$, and trajectory smoothness $J_p^*$) for each scenario and tested frequency as medians with interquartile ranges}
    \label{fig:newfreq_sensitivity}
\end{figure}

To compare the tested policy frequencies, we treated frequency as a categorical factor and analyzed the balanced $2\times3$ subset comprising both scenarios at 20, 40, and 60~Hz ($n=63$ navigation trips). Metrics were made dimensionless using the robot characteristic length $L_c=0.5$~m and the policy training time step duration as the characteristic time, $T_c=1/f_{\mathrm{train}}=1/20=0.05$~s., accordingly, minimum inner robot distance $d^* =d/L_c$, action smoothness $J_a^*=J_a * T_c^2$, and trajectory smoothness $J_p^*=J_p*T_c^3/L_c$. Using the fixed training timescale avoids introducing the tested execution frequency directly into the normalization. Figure~\ref{fig:newfreq_sensitivity} shows the dimensionless metrics for each scenario and tested frequency as medians with interquartile ranges.

A two-factor Freedman-Lane residual-permutation ANOVA tested frequency, scenario, and their interaction. Holm correction was applied separately to each effect family across the three primary metrics. Frequency significantly affected minimum inter-robot distance ($F(2,57)=4.59$, $p_{\rm perm}=0.0133$, $p_{\rm Holm}=0.0266$, $\eta_p^2=0.139$) and trajectory smoothness ($F(2,57)=14.47$, $p_{\rm perm}<0.001$, $p_{\rm Holm}<0.001$, $\eta_p^2=0.337$), but not action smoothness ($F(2,57)=2.23$, $p_{\rm Holm}=0.116$). Scenario affected minimum inner-robot distance ($F(1,57)=5.97$, $p_{\rm Holm}=0.0356$) and trajectory smoothness ($F(1,57)=7.37$, $p_{\rm Holm}=0.0336$), while no frequency$\times$scenario interaction was detected (all $p_{\rm Holm}\ge0.353$). Path length and mean neighbor distance showed no frequency effect and are omitted from the figure. 120~Hz runs were also conducted, but showed a large decrease in task performance metrics and thus were not sampled enough to be included in the factorial analyses. We further refer readers to the additional materials for more details, including performance metrics at the whole task level.

These results reveal a temporal component of the sim-to-real gap. Increasing the policy command update frequency improves performance in simulation, as seen by a shorter navigation path, while maintaining safe inner-robot distance with no collisions, leading to direct improvement in the overall machine tending task performance. However, the same trend does not transfer to the physical system. In particular, policy command updates above the sensing rate can repeatedly act on unchanged or asynchronously updated observations, while communication, state estimation, and low-level control introduce additional delays that are absent or idealized in simulation. Thus, command update frequency cannot be treated as an implementation parameter independent of the sensing and control stack when transferring learned policies to physical robots. Additional experiments investigating the sensitivity to command update frequency are available in the additional materials.

\begin{figure*}[!t]
    \centering
    \includegraphics[width=0.43\linewidth]{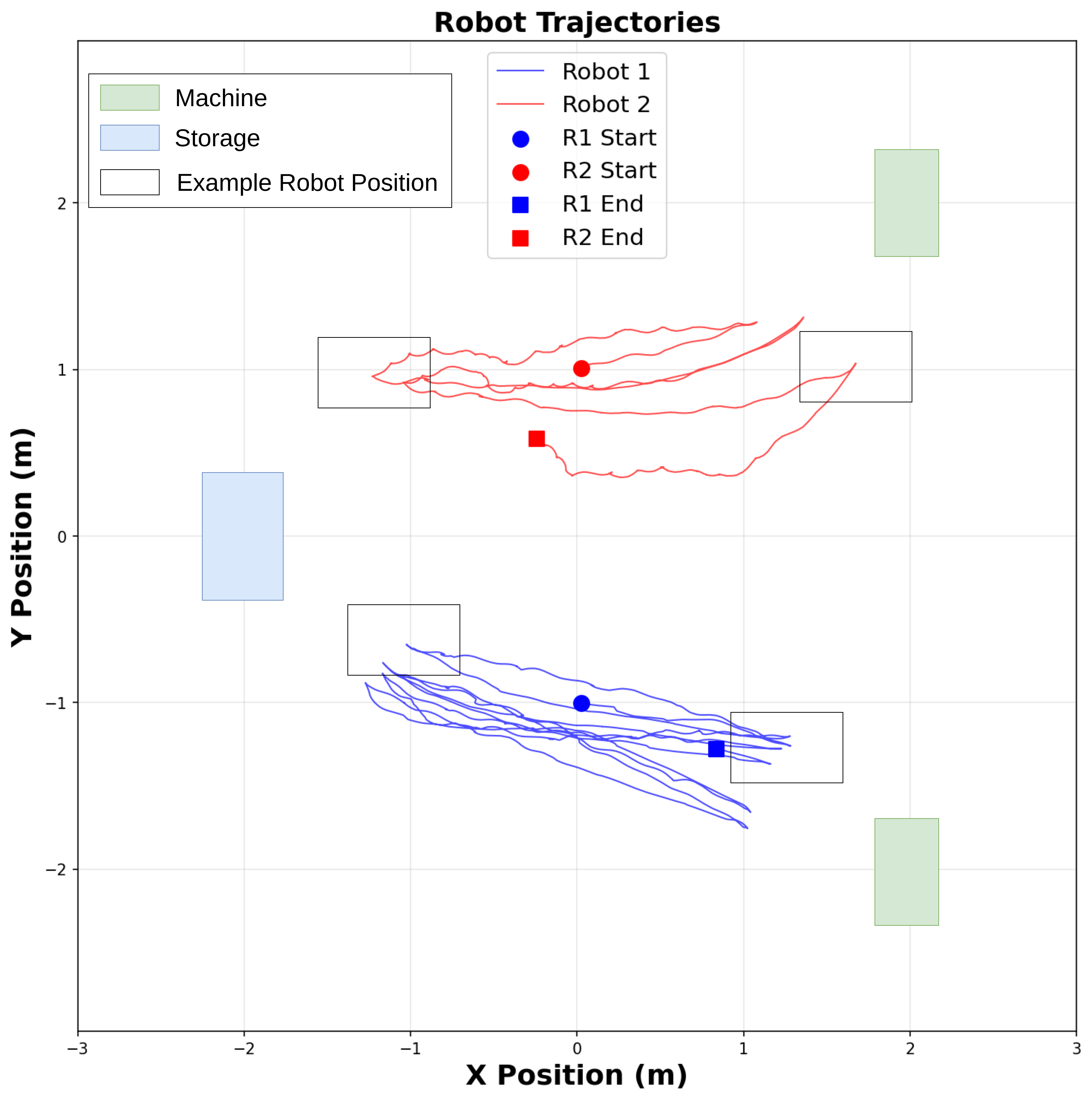}
    \includegraphics[width=0.435\linewidth]{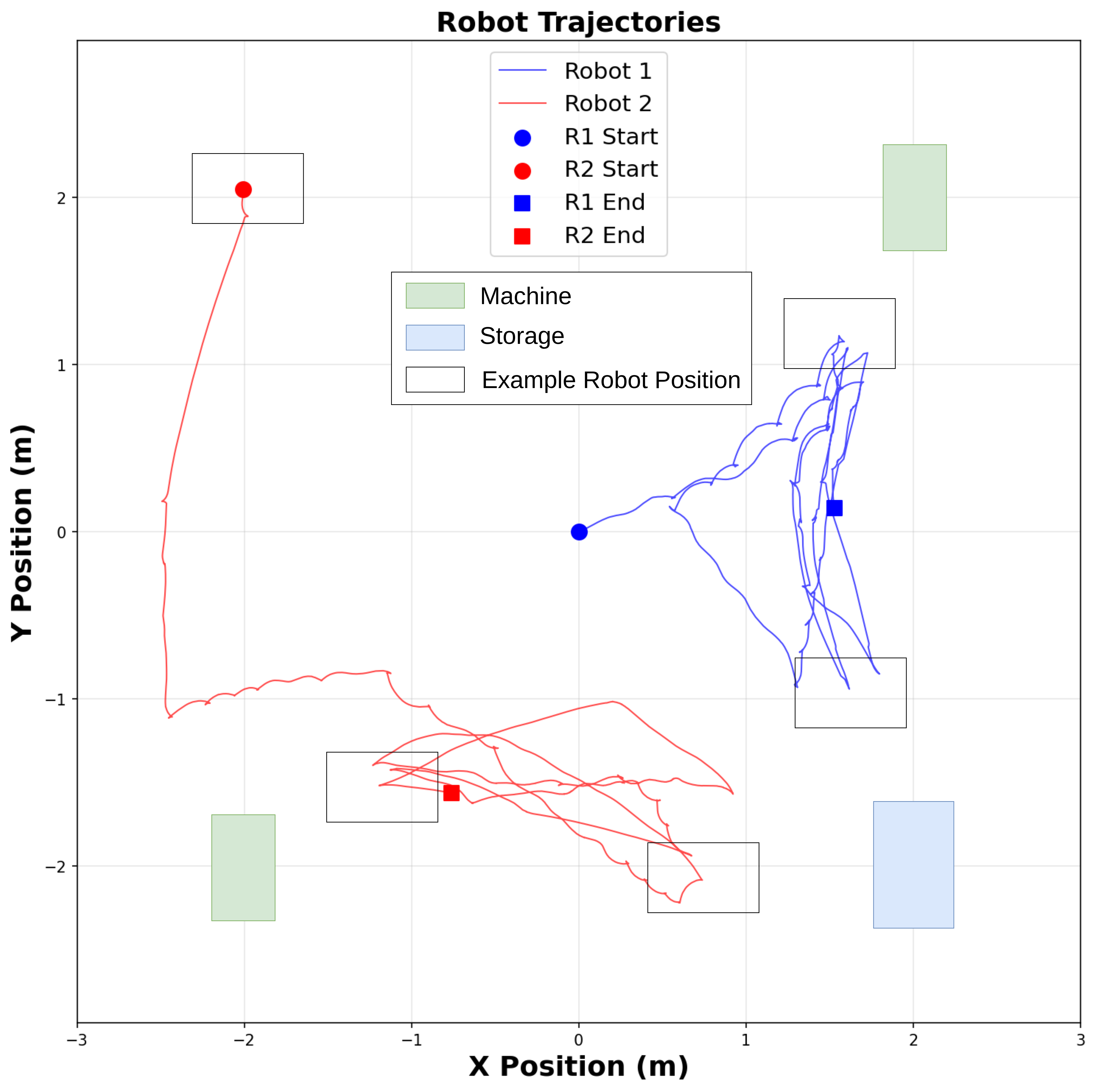}
    \caption{The full trajectories from the actual real-world deployment for scenario 1 and scenario 2 for the case of command update frequency of 40 Hz; showing the robots going for multiple trips of collection and delivery.}
    \label{fig:RealTrajectories}
\end{figure*}

Figure~\ref{fig:RealTrajectories} depicts the actual trajectories for the two robots in the real deployment for Scenario 1 (left side) and Scenario 2 (right side). Here we show the case where the model was tested with a command update frequency of 40 Hz. We can see that Scenario 2 is more challenging, requiring the two robots to navigate in proximity to each other because the dropping area is in the corner. We refer readers to check the videos for the real-world validation experiment \href{https://anonymouspapers123.github.io/FMAPPO/#Real}{online}.


\section{Conclusion}\label{sec:conc}
This paper investigated the application of decentralized multi-agent reinforcement learning for multi-robot multi-machine tending, addressing the challenges of coordination, task assignment, navigation, and safe operation in industrial setups. We introduced FMAPPO, a decentralized MARL framework that enables multiple mobile robots to collaboratively service multiple production machines. To the best of our knowledge, this is the first work to study MARL for combined task assignment and navigation for decentralized multi-robot multi-machine tending with actual physical mobile robots. The proposed model integrates 2D LiDAR measurements with task-related observations to learn a compact representation for decision-making. The experimental results demonstrate that the learned MARL policies can successfully zero-shot transfer from simulation to real robotic platforms while operating under realistic constraints, including onboard sensing, communication, and real-time onboard control. Moreover, this work also studies the sensitivity of the learned policy to command update frequency.

These results serve as a step toward the adoption of MARL-based solutions for real-world industrial applications. Future work can investigate including the object manipulation part and further improvements in robustness in more complex industrial settings.

\bibliographystyle{IEEEtran}
\bibliography{main}

\pagebreak

This part provides the additional materials for the paper Learning Multi-Agent Task Assignment and Navigation in the Factory: from Simulation to Real Robots. We still recommend that readers check the website\footnote{\url{https://anonymouspapers123.github.io/FMAPPO/\#Real}} for videos of the real-world validation experiments.

\section{Setup parameters}

\subsection{SMAPPO Implementation}
To evaluate against SMAPPO, we implemented its observation encoder using the same MAPPO implementation from the SKRL library~\cite{serrano2023skrl} that we used as the backbone for our FMAPPO implementation.

\begin{figure}[!t]
    \centering
    \includegraphics[width=1\linewidth]{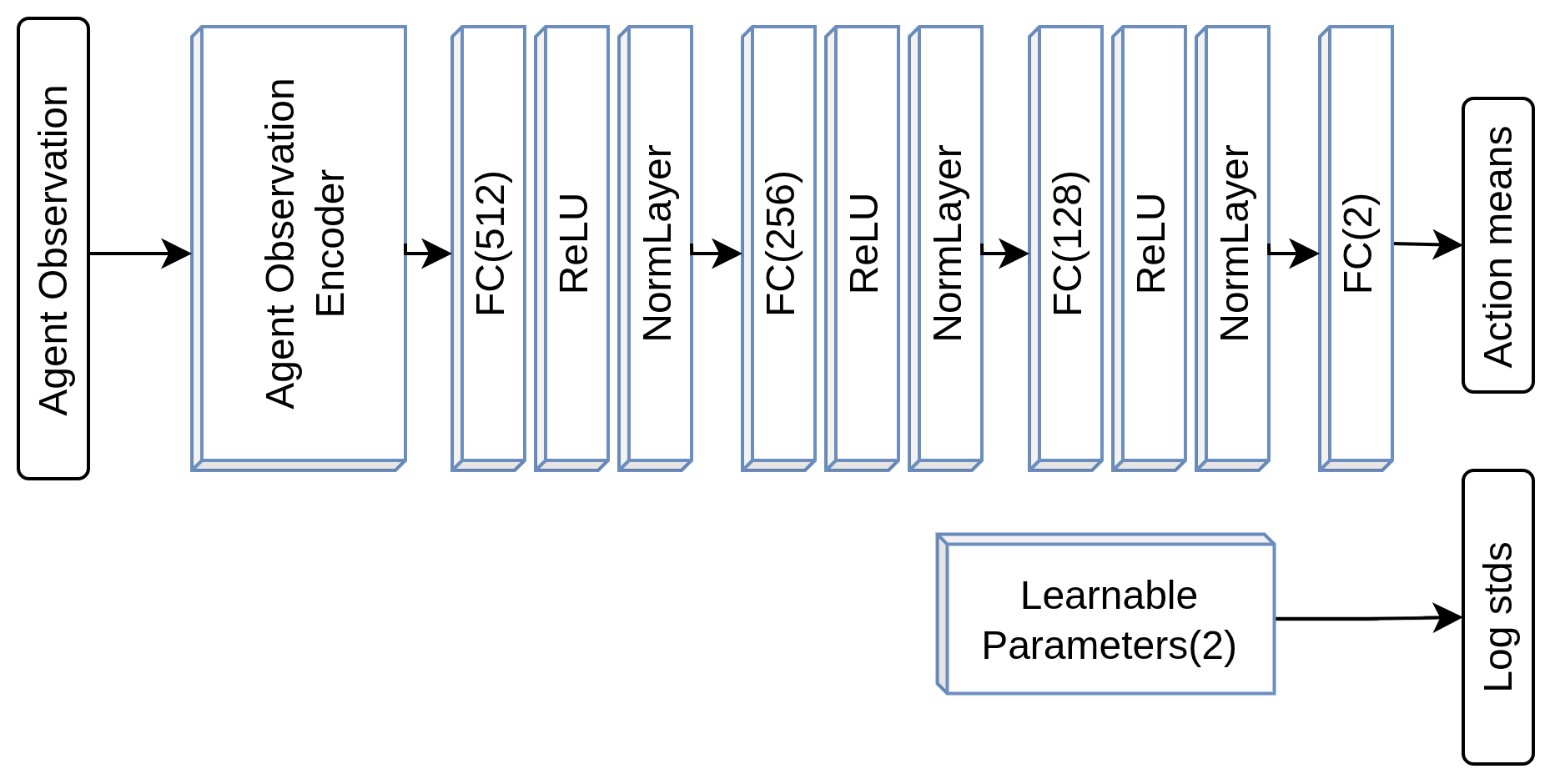}
    \caption{Full FMAPPO Actor Architecture: during training, the policy samples from a normal distribution using the predicted mean and log stds, while in evaluation mode the means are used directly as the action}
    \label{fig:FMAPPO_Actor}
\end{figure}

\begin{figure}[!t]
    \centering
    \includegraphics[width=1\linewidth]{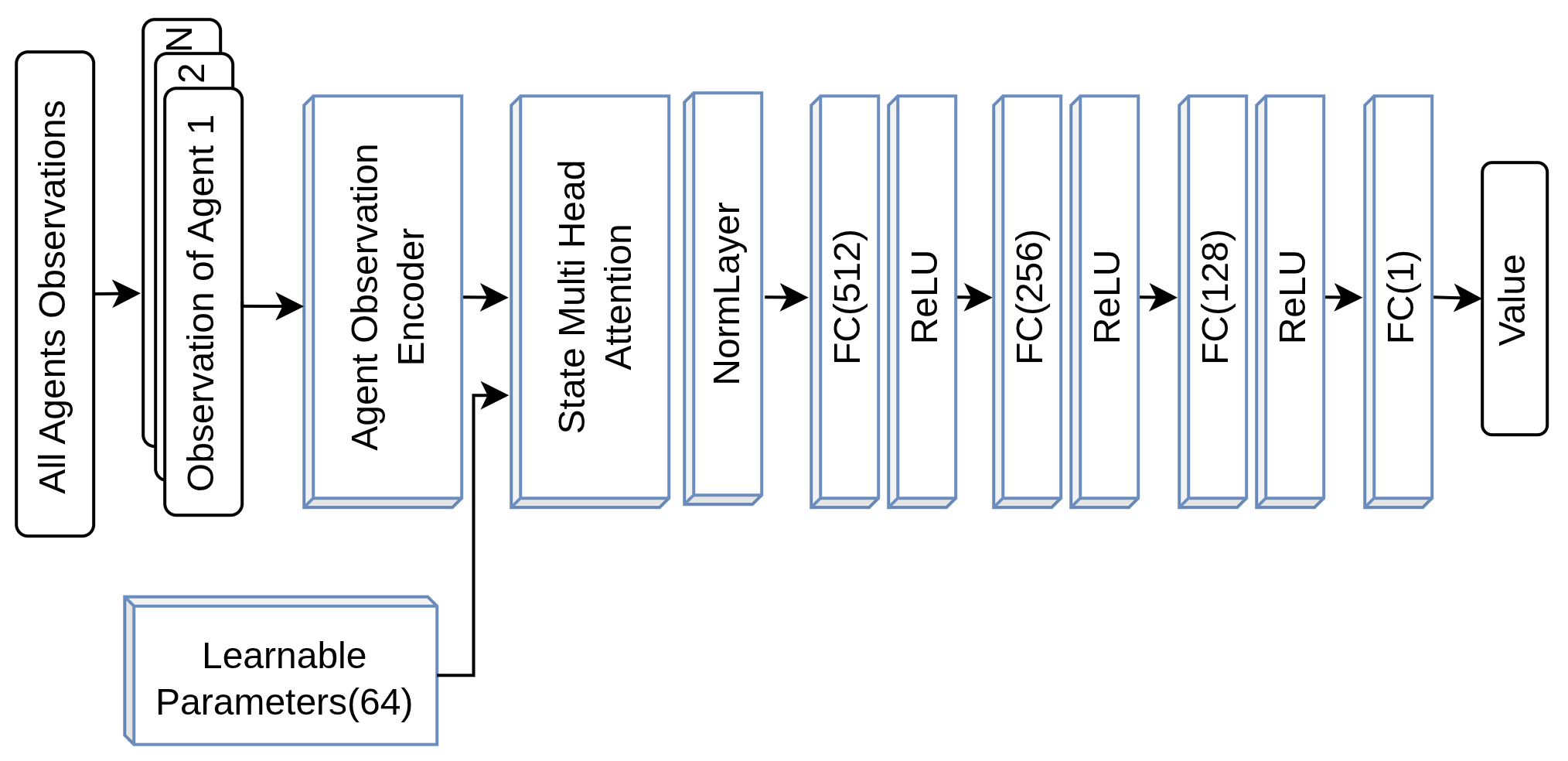}
    \caption{Full FMAPPO Critic Architecture with learnable parameters used as the query vector for the multi-head attention}
    \label{fig:FMAPPO_Critic}
\end{figure}

\begin{table}[ht]
\begin{center}
\caption{Model Parameters.}
\label{tab:Modelparams}
\begin{tabular}{|c|c|}
\hline
Parameter & Value  \\ \hline
\multicolumn{2}{|c|}{Lidar Processing Branch}  \\ \hline
LiDAR downsampled size & 270  (540 for the 2 lidars) \\ \hline
Conv1d layers kernal size & [3,3,3] \\ \hline
Conv1d layers channels num & [8,8,8] \\ \hline
MaxPool1d layers size & [4,4,4] \\ \hline
Lidar embed. dim. (FC size) & 8 \\ \hline

\multicolumn{2}{|c|}{Actor Parameters}  \\ \hline
Agent Enc. FC Size & 16  \\ \hline
Machines Enc. FC Size & 16  \\ \hline
Storages Enc. FC Size & 16  \\ \hline
Other Enc. FC Size & 16  \\ \hline
Atten. embed. dim. & 16  \\ \hline
Atten. heads & 2  \\ \hline

\multicolumn{2}{|c|}{Critic Parameters}  \\ \hline
Agent Enc. FC size & 18  \\ \hline
Machines Enc. FC size & 18  \\ \hline
Storages Enc. FC size & 18  \\ \hline
Other Enc. FC size & 18  \\ \hline
Atten. embed. dim. & 18  \\ \hline
Atten. heads & 3  \\ \hline
State Atten. embed. dim. & 64  \\ \hline

\multicolumn{2}{|c|}{Remaining Model Parameters}  \\ \hline
Rollouts & 24 \\ \hline
Learning epochs & 5 \\ \hline
Mini batches & 6 \\ \hline
Discount factor & 0.99 \\ \hline
Lambda & 0.95 \\ \hline
Optimizer & Adam \\ \hline
Learning rate & 3e-4 \\ \hline
Learning rate scheduler & KLAdaptiveRL \\ \hline
Learning rate scheduler kl\_threshold & 0.008 \\ \hline
Learning starts & 0 \\ \hline
Grad norm clip & 1.0 \\ \hline
Ratio clip & 0.2 \\ \hline
Value clip & 0.2 \\ \hline
Entropy loss scale & 0.001 \\ \hline
Value loss scale & 1.0 \\ \hline
State preprocessor & RunningStandardScaler \\ \hline
Shared\_state\_preprocessor & RunningStandardScaler \\ \hline
Value\_preprocessor & RunningStandardScaler \\ \hline
Rewards\_shaper & None \\ \hline
Time\_limit\_bootstrap & False \\ \hline
Mixed precision & False \\ \hline
Write interval & 180  \\ \hline
Checkpoint interval & 1800 \\ \hline
Base MAPPO implementation & Skrl Library   \\ \hline

\end{tabular}
\end{center}
\end{table}

\begin{table}[ht]
\begin{center}
\caption{Training Parameters.}
\label{tab:Trainparams}
\begin{tabular}{|c|c|}
\hline
Parameter & Value  \\ \hline
LiDAR range & 40 meters \\ \hline
Physics step size & 1/120 second \\ \hline
Decimation & 6 \\ \hline
Render Interval & 6 \\ \hline
Episode Length & 15 Seconds \\ \hline
Entites init. position & Random from uniform distribution \\ \hline
Entites init. position Range & (-2.1,+2.1) \\ \hline
Entites init. min. distance & robot radius + storage radius + 1.0 \\ \hline
LiDAR Sim. & MultiMeshRayCaster \\ \hline
LiDAR freq. in Sim. & 120 Hz \\ \hline
Num. Vertical channels & 1 \\ \hline
LiDAR Horizontal Range & [-135, 135] \\ \hline
LiDAR Horizontal resolution & 0.25 \\ \hline
Limit velocity & False \\ \hline
Reset on collision & False \\ \hline

Robot collision radius & 50 cm   \\ \hline
Machine collision radius & 44 cm   \\ \hline
Storage collision radius & 60 cm \\ \hline
Margin to collect or drop & 25 cm \\ \hline

\end{tabular}
\end{center}
\end{table}

\begin{table}[ht]
\begin{center}
\caption{Reward Parameters.}
\label{tab:Rewardparams}
\begin{tabular}{|c|c|}
\hline
Parameter & Value  \\ \hline
Collect reward & 15.0 \\ \hline
Deliver reward & 30.0 \\ \hline
Collision duration penalty & -1.0 \\ \hline
New collisions penalty & -25.0 \\ \hline
Uncollected part penalty & -0.1 \\ \hline
Dist reward scale & 20 \\ \hline
Time penalty & -0.01 \\ \hline

\end{tabular}
\end{center}
\end{table}

\section{Additional Results}
 In this section, we provide more insights into experiments described in the paper.

\subsection{Statistical Analysis of the Main Comparison Experiment}
Here we provide the statistical analysis of the results of the comparison experiment mentioned in the paper. Results are shown in Table~\ref{tab:statisticsMainExperiment}.

\begin{table*}[ht]
\centering
\caption{Main comparison experiment in the paper: Statistical Significance Testing FMAPPO vs MAPPO and SMAPPO (normality-selected test; effect size; Holm-Bonferroni-corrected p-values) }
\label{tab:statisticsMainExperiment}
\footnotesize
\begin{tabular}{|c|c|c|c|c|c|c|c|c|c|c|c|c|c|}
\hline
Model & Metric & M. Mean & Ours Mean & n & Diff & Shapiro-p & Normal & Test & p & p (Holm) & Sig (Holm) & Effect & E. Size \\ \hline
MAPPO & Collected & 2.50 & 3.71 & 7 & 1.22 & 0.327 & Yes & t-test & 6.1e-07 & 6.0e-06 & \textbf{Yes} & 8.24 & \textbf{large} \\ \hline
MAPPO & Delivered & 1.62 & 3.34 & 7 & 1.72 & 0.280 & Yes & t-test & 6.0e-07 & 6.0e-06 & \textbf{Yes} & 8.27 & \textbf{large} \\ \hline
MAPPO & Collisions & 2.06 & 1.69 & 7 & -0.36 & 0.107 & Yes & t-test & 0.002 & 0.009 & \textbf{Yes} & -2.03 & \textbf{large} \\ \hline
MAPPO & S. Score & 0.67 & 0.81 & 7 & 0.14 & 0.407 & Yes & t-test & 9.3e-06 & 6.5e-05 & \textbf{Yes} & 5.18 & \textbf{large} \\ \hline
MAPPO & Utilization & 0.62 & 0.93 & 7 & 0.30 & 0.327 & Yes & t-test & 6.1e-07 & 6.0e-06 & \textbf{Yes} & 8.24 & \textbf{large} \\ \hline
SMAPPO & Collected & 3.33 & 3.71 & 7 & 0.39 & 0.092 & Yes & t-test & 0.004 & 0.014 & \textbf{Yes} & 1.75 & \textbf{large} \\ \hline
SMAPPO & Delivered & 2.77 & 3.34 & 7 & 0.57 & 0.195 & Yes & t-test & 0.006 & 0.014 & \textbf{Yes} & 1.57 & \textbf{large} \\ \hline
SMAPPO & Collisions & 1.99 & 1.69 & 7 & -0.29 & 0.409 & Yes & t-test & 0.010 & 0.014 & \textbf{Yes} & -1.42 & \textbf{large} \\ \hline
SMAPPO & S. Score & 0.75 & 0.81 & 7 & 0.05 & 0.842 & Yes & t-test & 2.5e-04 & 0.002 & \textbf{Yes} & 2.91 & \textbf{large} \\ \hline
SMAPPO & Utilization & 0.83 & 0.93 & 7 & 0.10 & 0.092 & Yes & t-test & 0.004 & 0.014 & \textbf{Yes} & 1.75 & \textbf{large} \\ \hline
\end{tabular}
\end{table*}

\subsection{Command Update Frequency Simulation Experiment}
This section provides further results and analysis of the experiment reported in the paper about the effect of command update frequency change in simulation.

First, we start by providing the results of the statistical analysis described in the paper. Table~\ref{tab:sim_frequency_sensitivity_anova} shows the results of the permutation factorial ANOVA of frequency sensitivity (Holm--Bonferroni-corrected $p$-values on primary metrics; partial $\eta^{2}$ effect size).

\begin{table*}[ht]
\centering
\caption{Command frequency update experiment in simulation: Permutation factorial ANOVA of frequency sensitivity (Holm--Bonferroni-corrected $p$-values on primary metrics; partial $\eta^{2}$ effect size)}
\label{tab:sim_frequency_sensitivity_anova}
\footnotesize
\begin{tabular}{|c|c|c|c|c|c|c|c|c|}
\hline
Metric & Effect & $F$ & $\mathrm{df}_{e}$ & $\mathrm{df}_{\mathrm{err}}$ & $p$ & $\eta^{2}_{p}$ & $p$ (Holm) & Sig (Holm) \\ \hline
Min. distance & Frequency & 1.35 & 4 & 90 & 0.2602 & 0.057 & 0.5204 & No \\ \hline
Min. distance & Scenario & 824.40 & 1 & 90 & 0.0001 & 0.902 & 0.0004 & \textbf{Yes} \\ \hline
Min. distance & Freq.\ $\times$ Scen. & 0.97 & 4 & 90 & 0.4363 & 0.041 & 0.8726 & No \\ \hline
Position smoothness & Frequency & 0.08 & 4 & 90 & 0.9997 & 0.004 & 0.9997 & No \\ \hline
Position smoothness & Scenario & 259.71 & 1 & 90 & 0.0001 & 0.743 & 0.0004 & \textbf{Yes} \\ \hline
Position smoothness & Freq.\ $\times$ Scen. & 0.64 & 4 & 90 & 0.8089 & 0.028 & 0.8726 & No \\ \hline
Path length & Frequency & 2.84 & 4 & 90 & 0.0002 & 0.112 & 0.0008 & \textbf{Yes} \\ \hline
Path length & Scenario & 14.24 & 1 & 90 & 0.0001 & 0.137 & 0.0004 & \textbf{Yes} \\ \hline
Path length & Freq.\ $\times$ Scen. & 2.40 & 4 & 90 & 0.0010 & 0.096 & 0.0040 & \textbf{Yes} \\ \hline
Mean distance & Frequency & 3.56 & 4 & 90 & 0.0016 & 0.137 & 0.0048 & Yes \\ \hline
Mean distance & Scenario & 0.09 & 1 & 90 & 0.8080 & 0.001 & 0.8080 & No \\ \hline
Mean distance & Freq.\ $\times$ Scen. & 3.12 & 4 & 90 & 0.0034 & 0.122 & 0.0102 & \textbf{Yes} \\ \hline
\end{tabular}
\end{table*}

Table~\ref{tab:freq_significance} presents the statistical comparison of the 120Hz vs 10Hz performance (paired over 10 seeds; normality-selected test; effect size; Holm-Bonferroni-corrected p-values per scenario), where t-test refers to a paired t-test, Wilcox: Wilcoxon test, cost: Cost per part; Min. dist.: Min inter-robot distance; Min inter-robot distance; Mean dist.: Mean inter-robot distance.

\begin{table*}[ht]
\centering
\caption{Command frequency update experiment in simulation: Statistical comparison of 120Hz vs 10Hz performance (paired over 10 seeds; normality-selected test; effect size; Holm-Bonferroni-corrected p-values per scenario); t-test: paired t-test, Wilcox: Wilcoxon, cost: Cost per part; Min. dist.: Min inter-robot distance; Mean dist.: Mean inter-robot distance}
\label{tab:freq_significance}
\footnotesize
\begin{tabular}{|c|c|c|c|c|c|c|c|c|c|c|c|c|c|c|}
\hline
Scenario & Metric &  $\mu_{10Hz}$ & $\mu_{120Hz}$ & n & Diff & Shapiro-p & Normal & Test & p & p (Holm) & Sig & $Sig_{Holm}$ & Effect & E. Size \\ \hline
S1 & Collected & 14.00 & 14.00 & 10 & 0.00 & -- & -- & -- & 1.000 & 1.000 & No & No & 0.00 & neglig. \\ \hline
S1 & Delivered & 12.00 & 12.00 & 10 & 0.00 & -- & -- & -- & 1.000 & 1.000 & No & No & 0.00 & neglig. \\ \hline
S1 & Collisions & 0.00 & 0.00 & 10 & 0.00 & -- & -- & -- & 1.000 & 1.000 & No & No & 0.00 & neglig. \\ \hline
S1 & \textbf{Cost}& 2.35 & 2.31 & 10 & -0.05 & 0.542 & Yes & t-test & 6.2e-07 & 3.7e-06 & \textbf{Yes} & \textbf{Yes} & -3.89 & \textbf{large} \\ \hline
S1 & Min. dist. & 1.94 & 1.96 & 10 & 0.02 & 0.554 & Yes & t-test & 0.029 & 0.147 & \textbf{Yes} & No & 0.82 & large \\ \hline
S1 & Mean dist. & 2.57 & 2.58 & 10 & 0.01 & 0.670 & Yes & t-test & 0.588 & 1.000 & No & No & 0.18 & neglig. \\ \hline
S2 & \textbf{Collected} & 8.90 & 11.00 & 10 & 2.10 & 0.001 & No & Wilcox. & 0.007 & 0.026 & \textbf{Yes} & \textbf{Yes} & 1.00 & \textbf{large} \\ \hline
S2 & \textbf{Delivered} & 8.00 & 9.90 & 10 & 1.90 & 8.4e-05 & No & Wilcox. & 0.006 & 0.026 & \textbf{Yes} & \textbf{Yes} & 1.00 & \textbf{large} \\ \hline
S2 & Collisions & 0.40 & 0.00 & 10 & -0.40 & 4.7e-06 & No & Wilcox. & 0.157 & 0.315 & No & No & -1.00 & large \\ \hline
S2 & \textbf{Cost} & 3.13 & 2.48 & 10 & -0.65 & 2.7e-05 & No & Wilcox. & 0.002 & 0.012 & \textbf{Yes} & \textbf{Yes} & -1.00 & \textbf{large} \\ \hline
S2 & Min. dist. & 1.13 & 1.24 & 10 & 0.10 & 0.652 & Yes & t-test & 0.274 & 0.315 & No & No & 0.37 & small \\ \hline
S2 & \textbf{Mean dist.} & 2.40 & 2.65 & 10 & 0.25 & 1.3e-05 & No & Wilcox. & 0.002 & 0.012 & \textbf{Yes} & \textbf{Yes} & 1.00 & \textbf{large} \\ \hline
\end{tabular}
\end{table*}

\subsection{Ablation Study}

All models were trained for 500K environment interaction steps using 256 parallel environments, with the final 200 episodes used for evaluation, and each experiment was repeated seven times with different randomization seeds. Table~\ref{tab:statisticsAblation} shows the statistical significance testing of FMAPPO performance vs. the other models in the ablation study (normality-selected test; effect size; Holm-Bonferroni-corrected p-values with a significance level of $\alpha$=0.05). We also compared our model performance without all the added observation components to verify their combined effect and report the results in Table~\ref{tab:AblationAllperformance} and the statistical analysis of the results of this comparison in Table~\ref{tab:AblationAllstatistics}. We can see that the performance drops significantly with a large effect size in all evaluation metrics. In fact, this drop in performance is larger than the one created from just removing the LiDAR information (as in the ablation study in the main paper), indicating a significant combined benefit from including all of them.  

\begin{table*}[ht]
\centering
\caption{Ablation study on each component: Statistical Significance Testing vs FMAPPO (normality-selected test; effect size; Holm-Bonferroni-corrected p-values with a significance level of $\alpha$=0.05)}
\label{tab:statisticsAblation}
\footnotesize
\begin{tabular}{|c|c|c|c|c|c|c|c|c|c|c|c|c|c|c|}
\hline
Model & Metric & M. Mean & Ours Mean & n & Diff & Shapiro-p & Normal & Test & p & p(Holm) & Sig & Sig(Holm) & Effect & E. Size \\ \hline
NoLID & Collected & 3.60 & 3.71 & 7 & 0.12 & 0.788 & Yes & t-test & 0.002 & 0.037 & \textbf{Yes} & \textbf{Yes} & 1.88 & \textbf{large} \\ \hline
NoLID & Delivered & 3.17 & 3.34 & 7 & 0.17 & 0.668 & Yes & t-test & 0.007 & 0.090 & \textbf{Yes} & No & 1.49 & large \\ \hline
NoLID & Collisions & 2.04 & 1.69 & 7 & -0.35 & 0.492 & Yes & t-test & 0.012 & 0.133 & \textbf{Yes} & No & -1.34 & large \\ \hline
NoLID & S. Score & 0.77 & 0.81 & 7 & 0.04 & 0.493 & Yes & t-test & 0.005 & 0.070 & \textbf{Yes} & No & 1.61 & large \\ \hline
NoLID & Utilization & 0.90 & 0.93 & 7 & 0.03 & 0.788 & Yes & t-test & 0.002 & 0.037 & \textbf{Yes} & \textbf{Yes} & 1.88 & \textbf{large} \\ \hline
NoVel & Collected & 3.70 & 3.71 & 7 & 0.02 & 0.307 & Yes & t-test & 0.304 & 1.000 & No & No & 0.43 & small \\ \hline
NoVel & Delivered & 3.38 & 3.34 & 7 & -0.05 & 0.397 & Yes & t-test & 0.082 & 0.825 & No & No & -0.79 & medium \\ \hline
NoVel & Collisions & 1.91 & 1.69 & 7 & -0.21 & 0.638 & Yes & t-test & 0.090 & 0.825 & No & No & -0.76 & medium \\ \hline
NoVel & S. Score & 0.79 & 0.81 & 7 & 0.02 & 0.611 & Yes & t-test & 0.090 & 0.825 & No & No & 0.76 & medium \\ \hline
NoVel & Utilization & 0.92 & 0.93 & 7 & 3.8e-03 & 0.307 & Yes & t-test & 0.304 & 1.000 & No & No & 0.43 & small \\ \hline
NoOrient & Collected & 3.73 & 3.71 & 7 & -0.02 & 0.456 & Yes & t-test & 0.460 & 1.000 & No & No & -0.30 & small \\ \hline
NoOrient & Delivered & 3.37 & 3.34 & 7 & -0.04 & 0.818 & Yes & t-test & 0.473 & 1.000 & No & No & -0.29 & small \\ \hline
NoOrient & Collisions & 1.75 & 1.69 & 7 & -0.06 & 0.693 & Yes & t-test & 0.433 & 1.000 & No & No & -0.32 & small \\ \hline
NoOrient & S. Score & 0.80 & 0.81 & 7 & 4.0e-03 & 0.767 & Yes & t-test & 0.618 & 1.000 & No & No & 0.20 & neglig.\\ \hline
NoOrient & Utilization & 0.93 & 0.93 & 7 & -4.9e-03 & 0.456 & Yes & t-test & 0.460 & 1.000 & No & No & -0.30 & small \\ \hline
\end{tabular}
\end{table*}

\begin{table}[ht]
\centering
\caption{Ablation Study Final Performance Comparison (FMAPPO, NoLvOrient (FMAPPO with no LiDAR, Velocity, or Orientation information included)) mean (std)}
\label{tab:AblationAllperformance}
\begin{tabular}{|c|c|c|}
\hline
Metric & FMAPPO & NoLvOrient \\ \hline
Collected & \textbf{3.71 (0.05)} & 3.51 (0.06) \\ \hline 
Delivered & \textbf{3.34 (0.07)} & 3.00 (0.12) \\ \hline 
Collisions & \textbf{1.69 (0.15)} & 2.33 (0.42) \\ \hline 
Safety Score & \textbf{0.81 (0.02)} & 0.74 (0.04) \\ \hline 
Utilization & \textbf{0.93 (0.01)} & 0.88 (0.01) \\ \hline 
\end{tabular}
\end{table}

\begin{table*}[ht]
\centering
\caption{Ablation Study Statistical Significance Testing FMAPPO vs NoLvOrient (FMAPPO with no LiDAR, Velocity, or Orientation information included) (normality-selected test; effect size; Holm-Bonferroni-corrected p-values)}
\label{tab:AblationAllstatistics}
\footnotesize
\begin{tabular}{|c|c|c|c|c|c|c|c|c|c|c|c|c|c|c|}
\hline
Model & Metric & $\mu_{model}$ & $\mu_{Ours}$ & n & Diff & Shapiro-p & Normal & Test & p & p (Holm) & Sig & Sig (Holm) & Effect & E. Size \\ \hline
NoLvOrient & Collected & 3.51 & 3.71 & 7 & 0.20 & 0.381 & Yes & t-test & 3.8e-05 & 1.9e-04 & Yes & \textbf{Yes} & 4.07 & \textbf{large} \\ \hline
NoLvOrient & Delivered & 3.00 & 3.34 & 7 & 0.34 & 0.516 & Yes & t-test & 4.3e-04 & 0.001 & Yes & \textbf{Yes} & 2.63 & \textbf{large} \\ \hline
NoLvOrient & Collisions & 2.33 & 1.69 & 7 & -0.63 & 0.068 & Yes & t-test & 0.002 & 0.005 & Yes & \textbf{Yes} & -1.90 & \textbf{large} \\ \hline
NoLvOrient & S. Score & 0.74 & 0.81 & 7 & 0.07 & 0.050 & No & Wilcoxon & 0.016 & 0.016 & Yes & \textbf{Yes} & 1.00 & \textbf{large} \\ \hline
NoLvOrient & Utilization & 0.88 & 0.93 & 7 & 0.05 & 0.381 & Yes & t-test & 3.8e-05 & 1.9e-04 & Yes & \textbf{Yes} & 4.07 & \textbf{large} \\ \hline
\end{tabular}
\end{table*}

\subsection{Real Deployment Experiment}

\textbf{Frequency sensitivity analysis:} To study the effect of the policy command update frequency in the real deployment, we focused on the navigation aspect. Since each run involved multiple collections and deliveries by the two robots, we divided each run by the trips, measured navigation performance metrics for each trip, and then aggregated them over the entire run. We did not include the first collection trip because it starts from the initial robot position and not from the drop location. Also, we eliminated the first delivery trip because it can start on a slightly different side compared to the other trips.

Selected navigation metrics were minimum and mean inter-robot distance, action smoothness (measured from the jerk based on the second derivative of actions predicted by the policy), trajectory smoothness measured by the jerk computed from the third derivative of the robots' position logs during navigation, and path length of the trip. Then all metrics were made dimensionless as explained in the paper. Table~\ref{tab:frequency_sensitivity_anova} shows the statistical analysis results.

\begin{table*}[ht]
\centering
\caption{Real Deployment Experiment: Permutation factorial ANOVA of frequency sensitivity (Holm--Bonferroni-corrected $p$-values on primary metrics; partial $\eta^{2}$ effect size)}
\label{tab:frequency_sensitivity_anova}
\footnotesize
\begin{tabular}{|c|c|c|c|c|c|c|c|c|}
\hline
Metric & Effect & $F$ & $\mathrm{df}_{e}$ & $\mathrm{df}_{\mathrm{err}}$ & $p$ & $\eta^{2}_{p}$ & $p$ (Holm) & Sig (Holm) \\ \hline
Min. distance & Frequency & 4.59 & 2 & 57 & 0.0133 & 0.139 & 0.0266 & \textbf{Yes} \\ \hline
Min. distance & Scenario & 5.97 & 1 & 57 & 0.0178 & 0.095 & 0.0356 & \textbf{Yes} \\ \hline
Min. distance & Freq.\ $\times$ Scen. & 0.54 & 2 & 57 & 0.5811 & 0.019 & 1.0 & No \\ \hline
Action smoothness & Frequency & 2.23 & 2 & 57 & 0.1156 & 0.073 & 0.1156 & No \\ \hline
Action smoothness & Scenario & 0.25 & 1 & 57 & 0.6192 & 0.004 & 0.6192 & No \\ \hline
Action smoothness & Freq.\ $\times$ Scen. & 2.19 & 2 & 57 & 0.1178 & 0.071 & 0.3534 & No \\ \hline
Position smoothness & Frequency & 14.47 & 2 & 57 & 0.0001 & 0.337 & 0.0003 & \textbf{Yes} \\ \hline
Position smoothness & Scenario & 7.37 & 1 & 57 & 0.0112 & 0.114 & 0.0336 & \textbf{Yes} \\ \hline
Position smoothness & Freq.\ $\times$ Scen. & 0.03 & 2 & 57 & 0.976 & 0.001 & 1.0 & No \\ \hline
Path length & Frequency & 0.48 & 2 & 57 & 0.627 & 0.017 & -- & -- \\ \hline
Path length & Scenario & 1.53 & 1 & 57 & 0.2263 & 0.026 & -- & -- \\ \hline
Path length & Freq.\ $\times$ Scen. & 1.70 & 2 & 57 & 0.2039 & 0.056 & -- & -- \\ \hline
Mean distance & Frequency & 1.72 & 2 & 57 & 0.1859 & 0.057 & -- & -- \\ \hline
Mean distance & Scenario & 4.78 & 1 & 57 & 0.0336 & 0.077 & -- & -- \\ \hline
Mean distance & Freq.\ $\times$ Scen. & 0.53 & 2 & 57 & 0.5909 & 0.018 & -- & -- \\ \hline
\end{tabular}
\end{table*}

\begin{figure}[!t]
    \centering
    \includegraphics[width=0.75\linewidth]{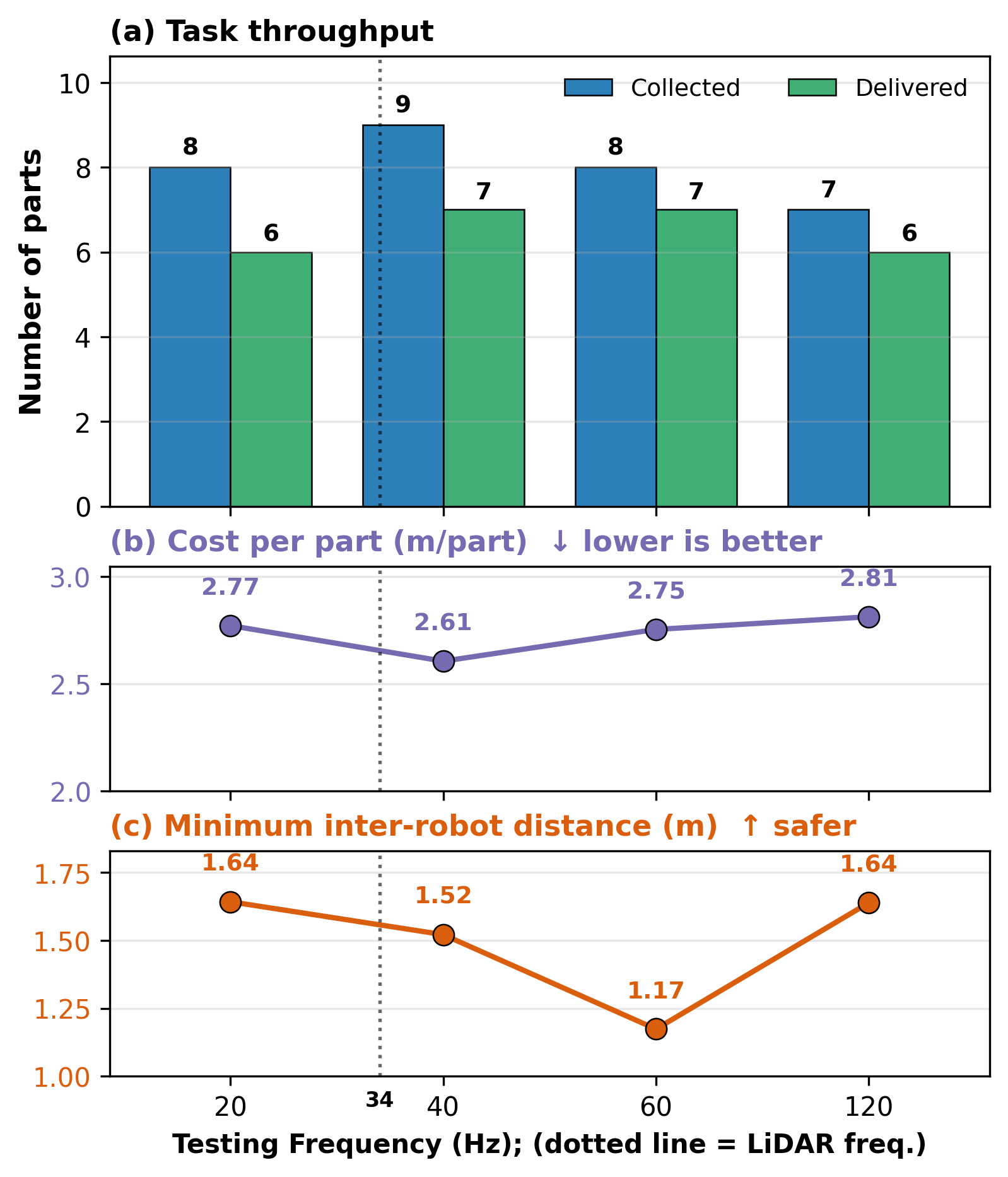}
    \caption{Real Deployment first layout results: showing (a) task throughput (total collected parts, total delivered parts), (b) efficiency as collection cost, and (c) minimum distance between the two robots.}
    \label{fig:Scenario1_results}
\end{figure}

\begin{figure}[!t]
    \centering
    \includegraphics[width=0.75\linewidth]{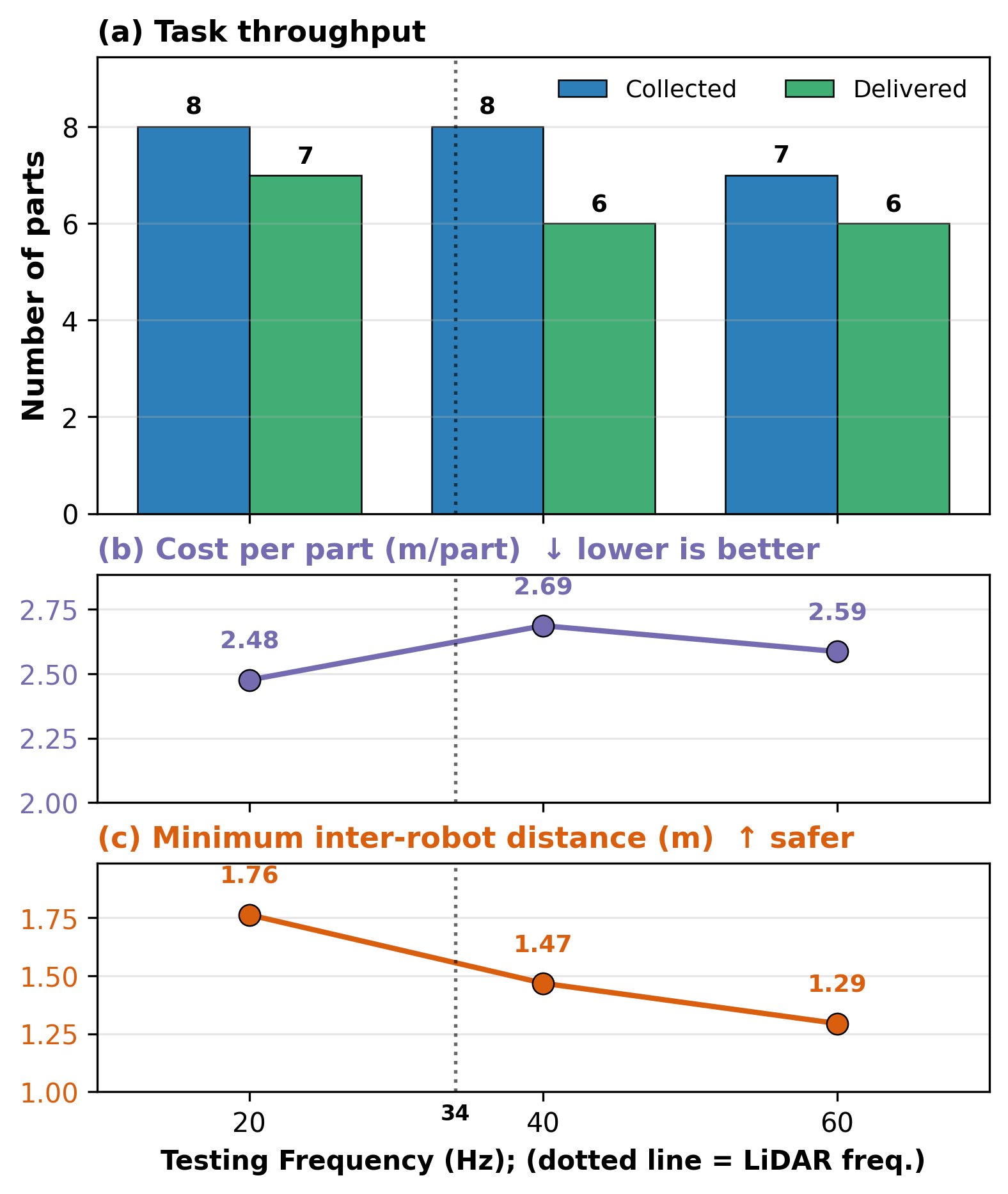}
    \caption{Real Deployment second layout results: showing (a) task throughput (total collected parts, total delivered parts), (b) efficiency as collection cost, and (c) minimum distance between the two robots.}
    \label{fig:Scenario2_results}
\end{figure}

\textbf{Overall task Performance:} During the real deployment experiment, we also monitored the total number of collected parts, delivered parts, and cost per part, as ($CC =  TraveledDistance_{total}/ (Collected+Delivered)$), and the minimum inner-robot distance at each single frequency run. Figures~\ref{fig:Scenario1_results} and \ref{fig:Scenario2_results} show the results for scenario one and two, respectively.

In the first scenario, the model performed the task successfully and safely with no collisions at the different tested frequencies, except for a minor side touch for one box at 120~Hz. It showed an increase in throughput and a drop in cost when doubling the command update frequency, but it goes down as the testing frequency increases. In terms of safety, the minimum distance between the two robots goes down as the frequency of command updates increases, except at 120 Hz, where the distance goes up again, but at the cost of reducing throughput. Even though, at the large task (machine tending) level, it is a single run per frequency and can not be used to establish a statistical trend, it was useful for initial insights and to remove the 120 Hz frequency from further tests.

In the second scenario, again, the model successfully and safely performed the task without collision and showed a slight drop in throughput and minimum distance with an increase in the collection cost as the testing command update frequency increased more and more from the original training frequency.

\section{Additional Experiments}
\subsection{Sample Efficiency}

\begin{figure}
     \centering
     \includegraphics[width=1\linewidth]{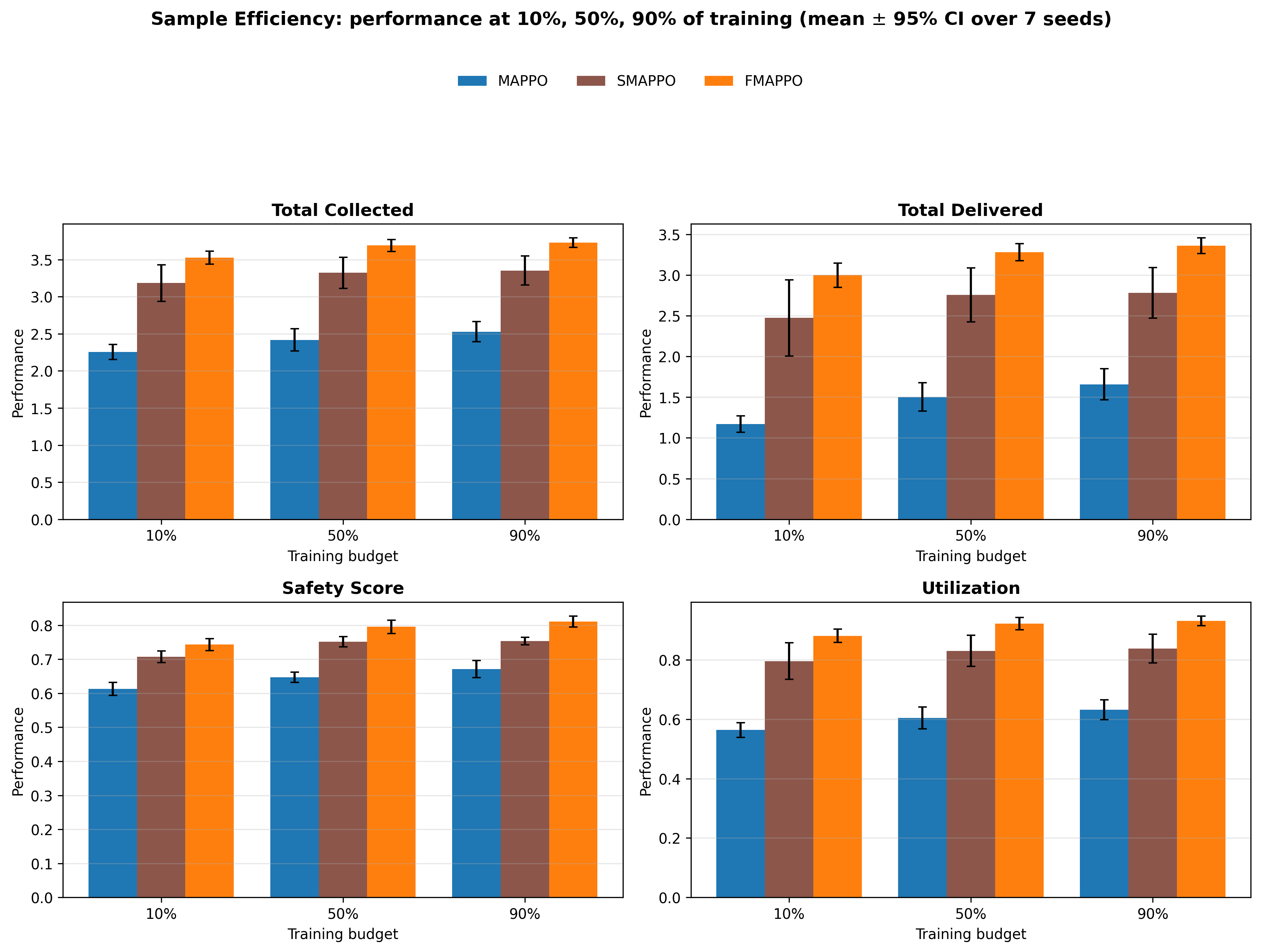}
     \caption{Sample efficiency: FMAPPO vs MAPPO and SMAPPO achieved performance at 10\%, 50\%, 90\% of the training budget, mean (std) over 7 seeds (per-seed trailing 10-episode window) with\textbf{ fixed part processing} time of 150 steps}
     \label{fig:sample_efficiencyMain}
 \end{figure}

Here we compare FMAPPO against MAPPO and SMAPPO in terms of sample efficiency, given a limited budget of training. So we compared the performance of the three models with 10\%, 50\%, and 90\% of the full training time and report the mean and standard deviation over seven seeds in Figure~\ref{fig:sample_efficiencyMain}, and the statistical significance of the results is analyzed in Table~\ref{tab:sample_efficiency_statsMain}.

\begin{table*}[ht]
\centering
\caption{Sample efficiency: Statistical significance of performance at 50\% of training vs FMAPPO (normality-selected test; effect size; Holm-Bonferroni-corrected p-values) with\textbf{ fixed part processing} time of 150 steps}
\label{tab:sample_efficiency_statsMain}
\footnotesize
\begin{tabular}{|c|c|c|c|c|c|c|c|c|c|c|c|c|c|c|}
\hline
Model & Metric & $\mu_{model}$ & $\mu_{ours}$ & n & Diff & Shapiro-p & Normal & Test & p & $p_{Holm}$ & Sig & $Sig_{Holm}$ & Effect & E. Size \\ \hline
MAPPO & Collected & 2.42 & 3.69 & 7 & 1.27 & 0.394 & Yes & t-test & 6.2e-07 & 4.3e-06 & Yes & \textbf{Yes} & 8.22 & \textbf{large} \\ \hline
MAPPO & Delivered & 1.50 & 3.28 & 7 & 1.78 & 0.630 & Yes & t-test & 4.8e-07 & 3.8e-06 & Yes & \textbf{Yes} & 8.58 & \textbf{large} \\ \hline
MAPPO & S. Score & 0.65 & 0.80 & 7 & 0.15 & 0.987 & Yes & t-test & 2.3e-05 & 1.2e-04 & Yes & \textbf{Yes} & 4.43 & \textbf{large} \\ \hline
MAPPO & Utilization & 0.60 & 0.92 & 7 & 0.32 & 0.394 & Yes & t-test & 6.2e-07 & 4.3e-06 & Yes & \textbf{Yes} & 8.22 & \textbf{large} \\ \hline
SMAPPO & Collected & 3.32 & 3.69 & 7 & 0.37 & 0.143 & Yes & t-test & 0.009 & 0.028 & Yes & \textbf{Yes} & 1.42 & \textbf{large} \\ \hline
SMAPPO & Delivered & 2.76 & 3.28 & 7 & 0.53 & 0.067 & Yes & t-test & 0.013 & 0.028 & Yes & \textbf{Yes} & 1.33 & \textbf{large} \\ \hline
SMAPPO & S. Score & 0.75 & 0.80 & 7 & 0.04 & 0.211 & Yes & t-test & 0.003 & 0.012 & Yes & \textbf{Yes} & 1.81 & \textbf{large} \\ \hline
SMAPPO & Utilization & 0.83 & 0.92 & 7 & 0.09 & 0.143 & Yes & t-test & 0.009 & 0.028 & Yes & \textbf{Yes} & 1.42 & \textbf{large} \\ \hline
\end{tabular}
\end{table*}

\subsection{Random Part Processing Time}
\begin{figure*}
    \centering
    \includegraphics[width=1\linewidth]{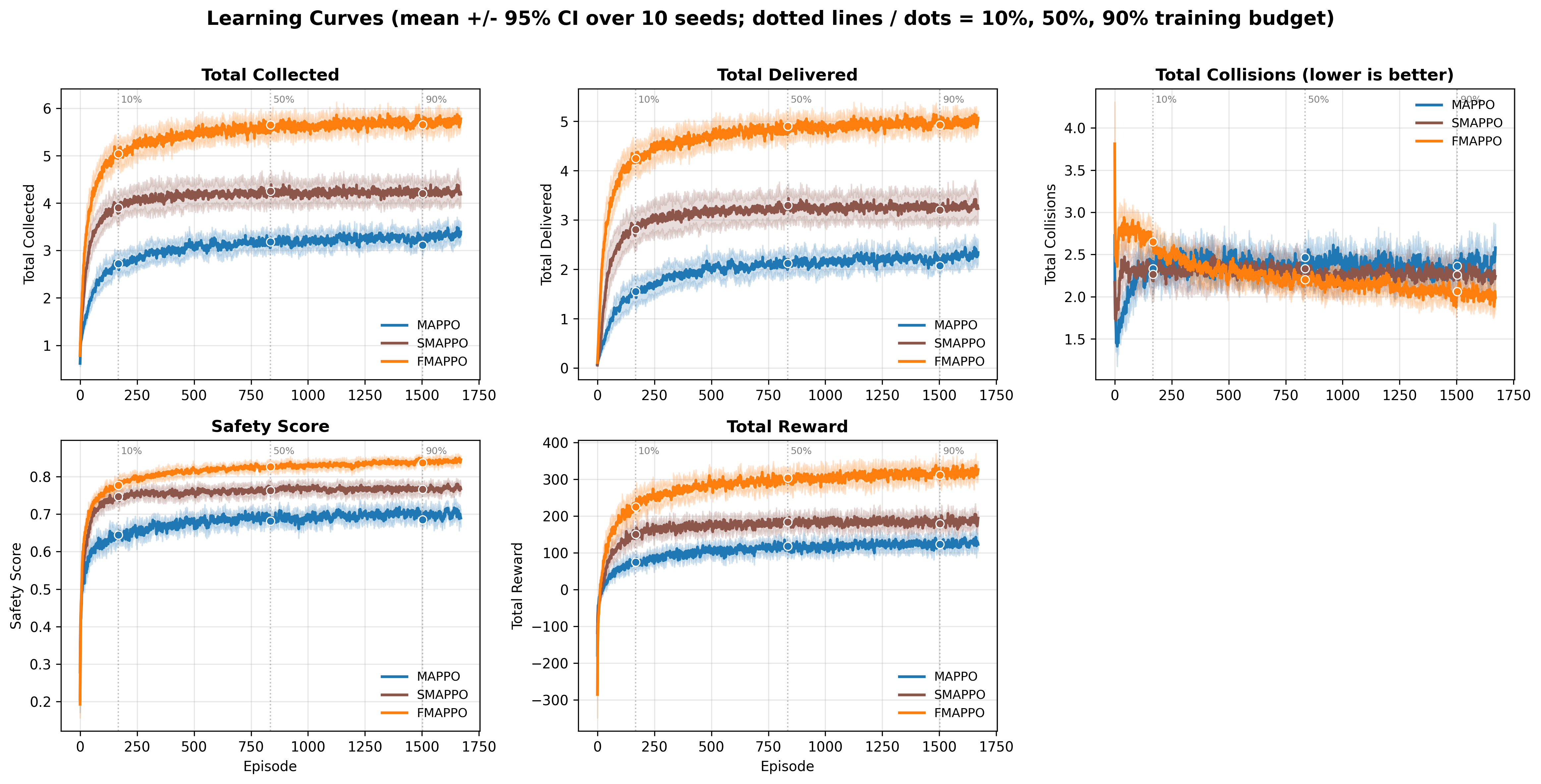}
    \caption{Random Part Processing Time Learning curves overview over 10 seeds}
    \label{fig:RandomLearningCurves}
\end{figure*}

\begin{figure}
    \centering
    \includegraphics[width=1\linewidth]{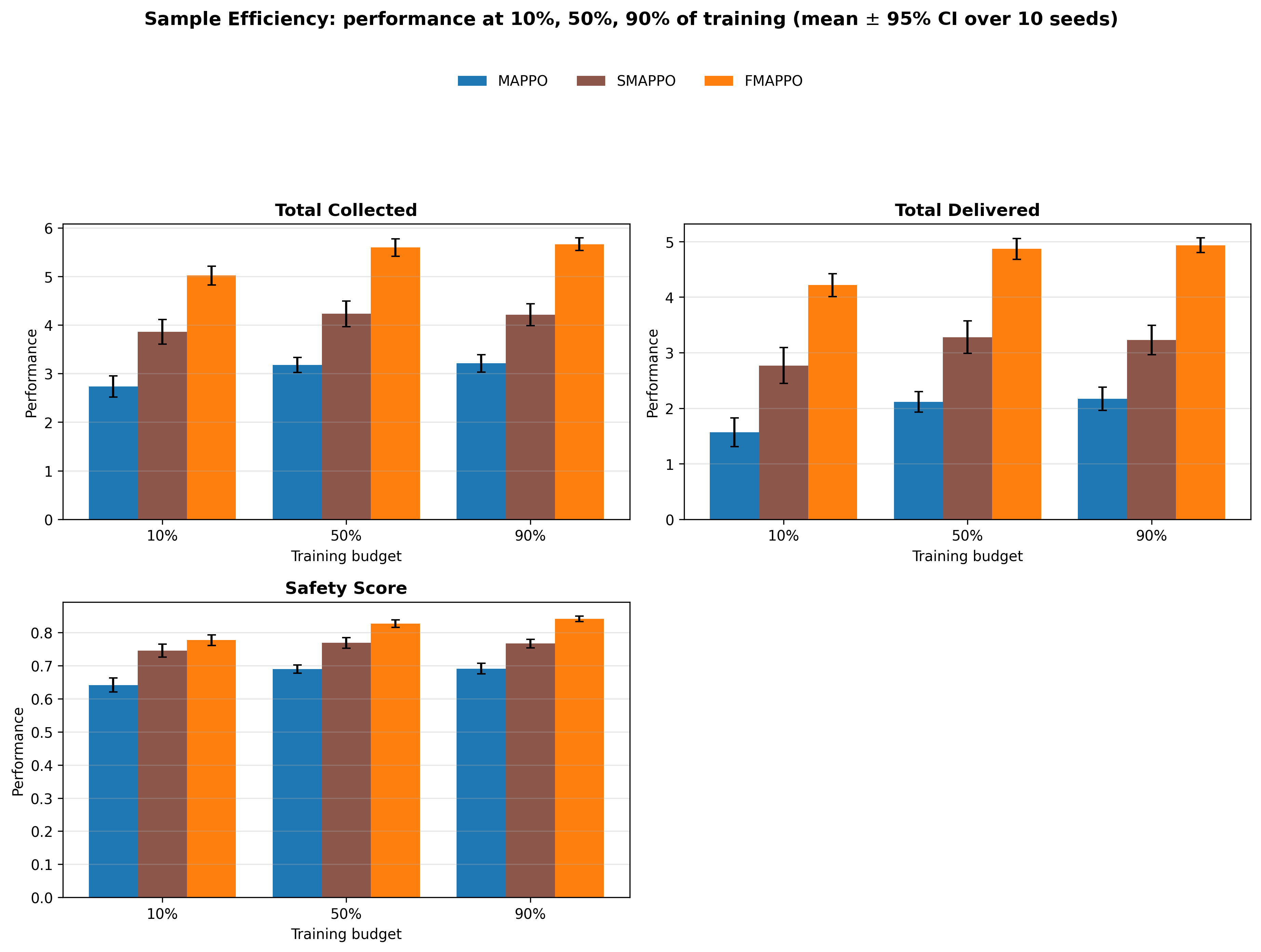}
    \caption{Sample efficiency: achieved performance at 10\%, 50\%, 90\% of the training budget, mean (std) over 10 seeds (per-seed trailing 10-episode window) with \textbf{random part processing time} between 60 and 900 environment steps}
    \label{fig:sample_efficiency_statsRandom}
\end{figure}

Here we compare FMAPPO with MAPPO under random part processing time. This aims to show the robustness of FMAPPO under random processing delays. The processing time of each part is randomly sampled from a uniform distribution between 60 and 900 environment steps. Each model is trained for 500K environment steps, with the results shown below. The models are trained with the LiDAR frequency limited to 34Hz to better reflect the real-world deployment.
In this experiment, the utilization metric can not be calculated under random processing time, since the maximum number of parts that can be processed per episode is not fixed, and the utilization metric is defined based on that. Table~\ref{tab:RandomPartsperformance} reports the results, while Figure~\ref{fig:RandomLearningCurves} presents the learning curves and Table~\ref{tab:RandomPartsstatistics} reports the statistical analysis with normality-selected test, effect size, and Holm-Bonferroni-corrected p-values, and Figure~\ref{fig:sample_efficiency_statsRandom} presents the sample efficiency; the statistical analysis of the significance of the sample efficiency is reported in Table~\ref{tab:sample_efficiency_statsRandom}. FMAPPO outperforms both SMAPPO and MAPPO with a statistically significant large effect size in parts collection, delivery, and safety score.

\begin{table*}[ht]
\centering
\caption{Random Part Processing Time: Final Performance Comparison (FMAPPO, MAPPO, SMAPPO) mean (std)}
\label{tab:RandomPartsperformance}
\begin{tabular}{|c|c|c|c|c|c|}
\hline
Metric & FMAPPO & MAPPO & Imp. vs MAPPO & SMAPPO & Imp. vs SMAPPO \\ \hline
Collected & \textbf{5.71 (0.18)} & 3.30 (0.23) & 73.03\% & 4.23 (0.32) & 34.99\% \\ \hline 
Delivered & \textbf{4.98 (0.19)} & 2.26 (0.26) & 120.35\% & 3.26 (0.38) & 52.76\% \\ \hline 
Collisions & \textbf{2.02 (0.14)} & 2.39 (0.17) & $-15.48\%$ & 2.26 (0.19) & $-10.62\%$ \\ \hline 
Safety Score & \textbf{0.84 (0.01)} & 0.70 (0.02) & 0.14 & 0.77 (0.02) & 0.07 \\ \hline 
\end{tabular}
\end{table*}

\begin{table*}[ht]
\centering
\caption{Random parts processing time: Statistical Significance Testing vs FMAPPO (normality-selected test; effect size; Holm-Bonferroni-corrected p-values)}
\label{tab:RandomPartsstatistics}
\footnotesize
\begin{tabular}{|c|c|c|c|c|c|c|c|c|c|c|c|c|c|c|}
\hline
Model & Metric & $\mu_{model}$ & $\mu_{ours}$ & n & Diff & Shapiro-p & Normal & Test & p & p (Holm) & Sig & Sig (Holm) & Effect & E. Size \\ \hline
MAPPO & Collected & 3.30 & 5.71 & 10 & 2.41 & 0.257 & Yes & t-test & 5.6e-09 & 3.4e-08 & Yes & \textbf{Yes} & 6.68 & \textbf{large} \\ \hline
MAPPO & Delivered & 2.26 & 4.98 & 10 & 2.72 & 0.251 & Yes & t-test & 4.2e-09 & 3.4e-08 & Yes & \textbf{Yes} & 6.89 & \textbf{large} \\ \hline
MAPPO & Collisions & 2.39 & 2.02 & 10 & -0.37 & 0.228 & Yes & t-test & 4.4e-04 & 8.8e-04 & Yes & \textbf{Yes} & -1.70 & \textbf{large} \\ \hline
MAPPO & S. Score & 0.70 & 0.84 & 10 & 0.14 & 0.377 & Yes & t-test & 4.2e-09 & 3.4e-08 & Yes & \textbf{Yes} & 6.90 & \textbf{large} \\ \hline
SMAPPO & Collected & 4.23 & 5.71 & 10 & 1.48 & 0.520 & Yes & t-test & 1.2e-06 & 3.8e-06 & Yes & \textbf{Yes} & 3.62 & \textbf{large} \\ \hline
SMAPPO & Delivered & 3.26 & 4.98 & 10 & 1.72 & 0.390 & Yes & t-test & 9.6e-07 & 3.8e-06 & Yes & \textbf{Yes} & 3.70 & \textbf{large} \\ \hline
SMAPPO & Collisions & 2.26 & 2.02 & 10 & -0.23 & 0.855 & Yes & t-test & 0.009 & 0.009 & Yes & \textbf{Yes} & -1.05 & \textbf{large} \\ \hline
SMAPPO & S. Score & 0.77 & 0.84 & 10 & 0.07 & 0.634 & Yes & t-test & 7.1e-07 & 3.5e-06 & Yes & \textbf{Yes} & 3.83 & \textbf{large} \\ \hline
\end{tabular}
\end{table*}

\begin{table*}[ht]
\centering
\caption{Random parts processing time: Statistical significance of performance at 50\% of training vs FMAPPO (normality-selected test; effect size; Holm-Bonferroni-corrected p-values)}
\label{tab:sample_efficiency_statsRandom}
\footnotesize
\begin{tabular}{|c|c|c|c|c|c|c|c|c|c|c|c|c|c|c|}
\hline
Model & Metric & $\mu_{model}$ & $\mu_{ours}$ & n & Diff & Shapiro-p & Normal & Test & p & p (Holm) & Sig & Sig (Holm) & Effect & E. Size \\ \hline
MAPPO & Collected & 3.18 & 5.60 & 10 & 2.42 & 0.748 & Yes & t-test & 2.2e-09 & 1.1e-08 & Yes & \textbf{Yes} & 7.43 & \textbf{large} \\ \hline
MAPPO & Delivered & 2.11 & 4.87 & 10 & 2.76 & 0.787 & Yes & t-test & 1.8e-09 & 1.1e-08 & Yes & \textbf{Yes} & 7.61 & \textbf{large} \\ \hline
MAPPO & S. Score & 0.69 & 0.83 & 10 & 0.14 & 0.868 & Yes & t-test & 1.0e-08 & 4.1e-08 & Yes & \textbf{Yes} & 6.24 & \textbf{large} \\ \hline
SMAPPO & Collected & 4.23 & 5.60 & 10 & 1.37 & 0.928 & Yes & t-test & 1.8e-06 & 3.6e-06 & Yes & \textbf{Yes} & 3.44 & \textbf{large} \\ \hline
SMAPPO & Delivered & 3.28 & 4.87 & 10 & 1.59 & 0.989 & Yes & t-test & 8.2e-07 & 2.5e-06 & Yes & \textbf{Yes} & 3.77 & \textbf{large} \\ \hline
SMAPPO & S. Score & 0.77 & 0.83 & 10 & 0.06 & 0.662 & Yes & t-test & 7.1e-06 & 7.1e-06 & Yes & \textbf{Yes} & 2.91 &\textbf{large} \\ \hline
\end{tabular}
\end{table*}

\subsection{More Command Update Frequency Experiments}
\begin{figure*}[!t]
    \centering
    \includegraphics[width=0.499\linewidth]{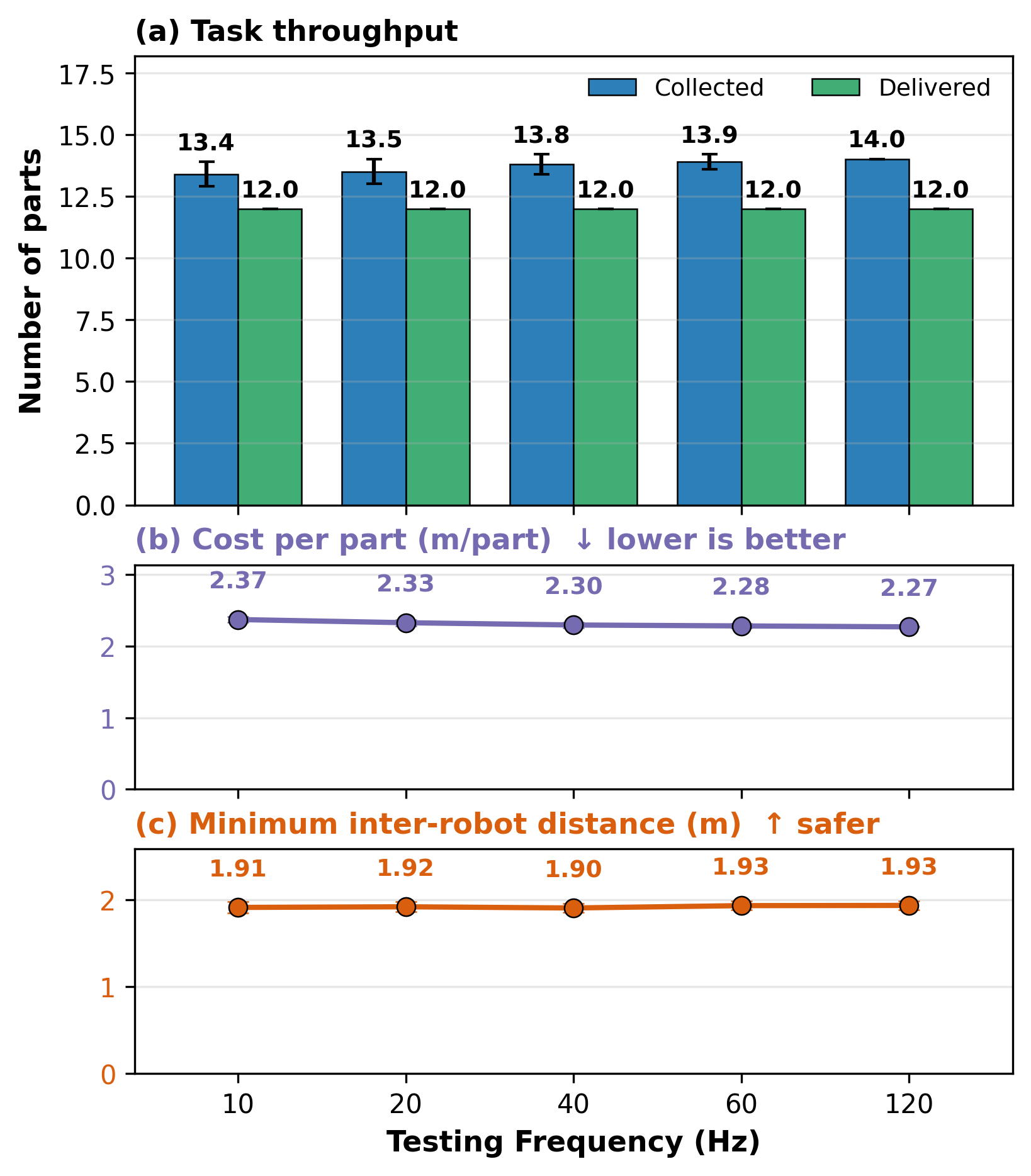}
    \includegraphics[width=0.485\linewidth]{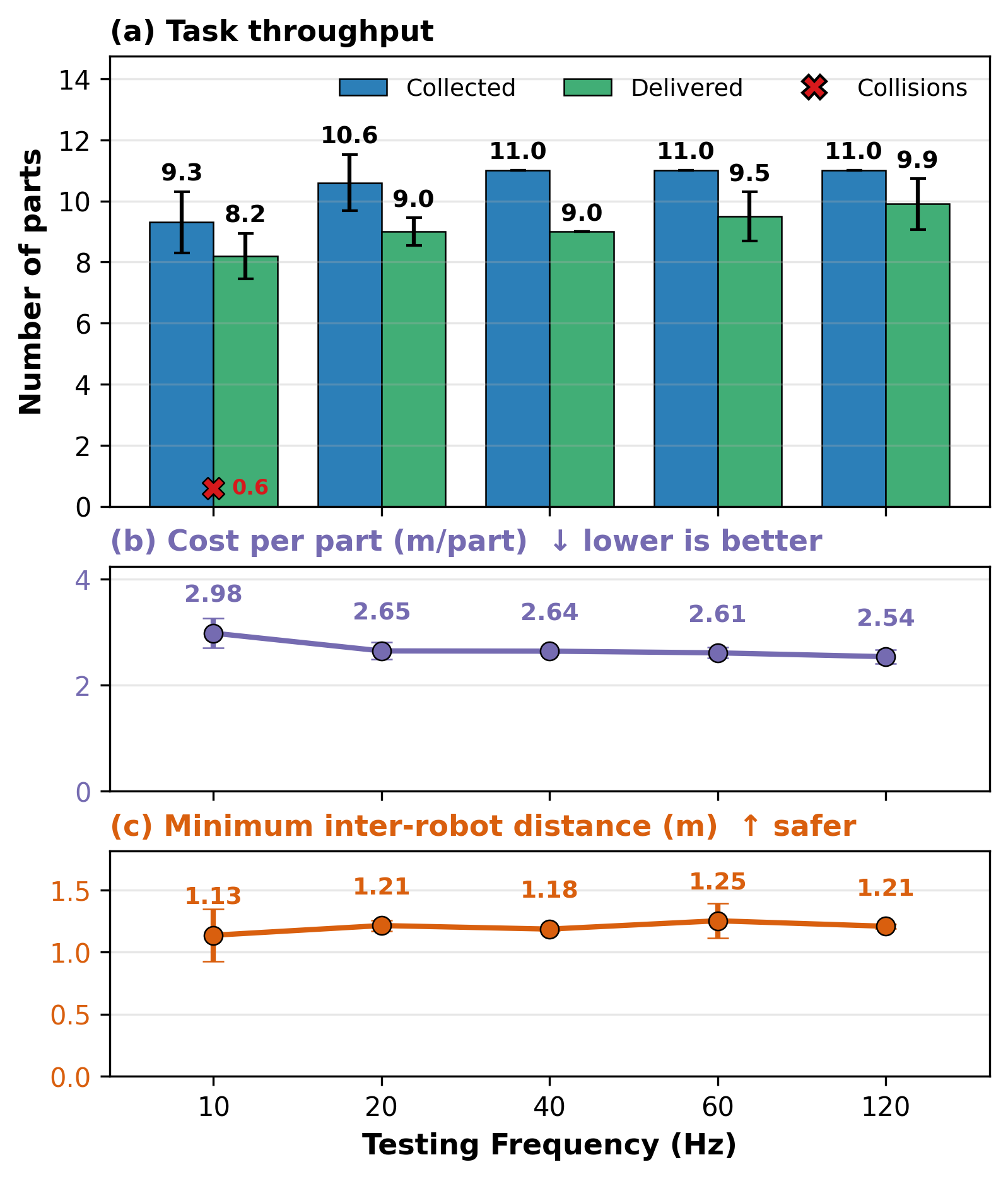}

    \caption{Position ($\sigma$ 2 cm), velocity ($\sigma$ 2 cm/s), LiDAR ($\sigma$ 1 cm), action ($\sigma$ 1 cm/s), and initial positions ($\sigma$ 10 cm), in the left Scenario 1, and in the right Scenario 2}
    \label{fig:freqP1_O02L01A01}
\end{figure*}

\begin{figure*}[!t]
    \centering
    \includegraphics[width=0.499\linewidth]{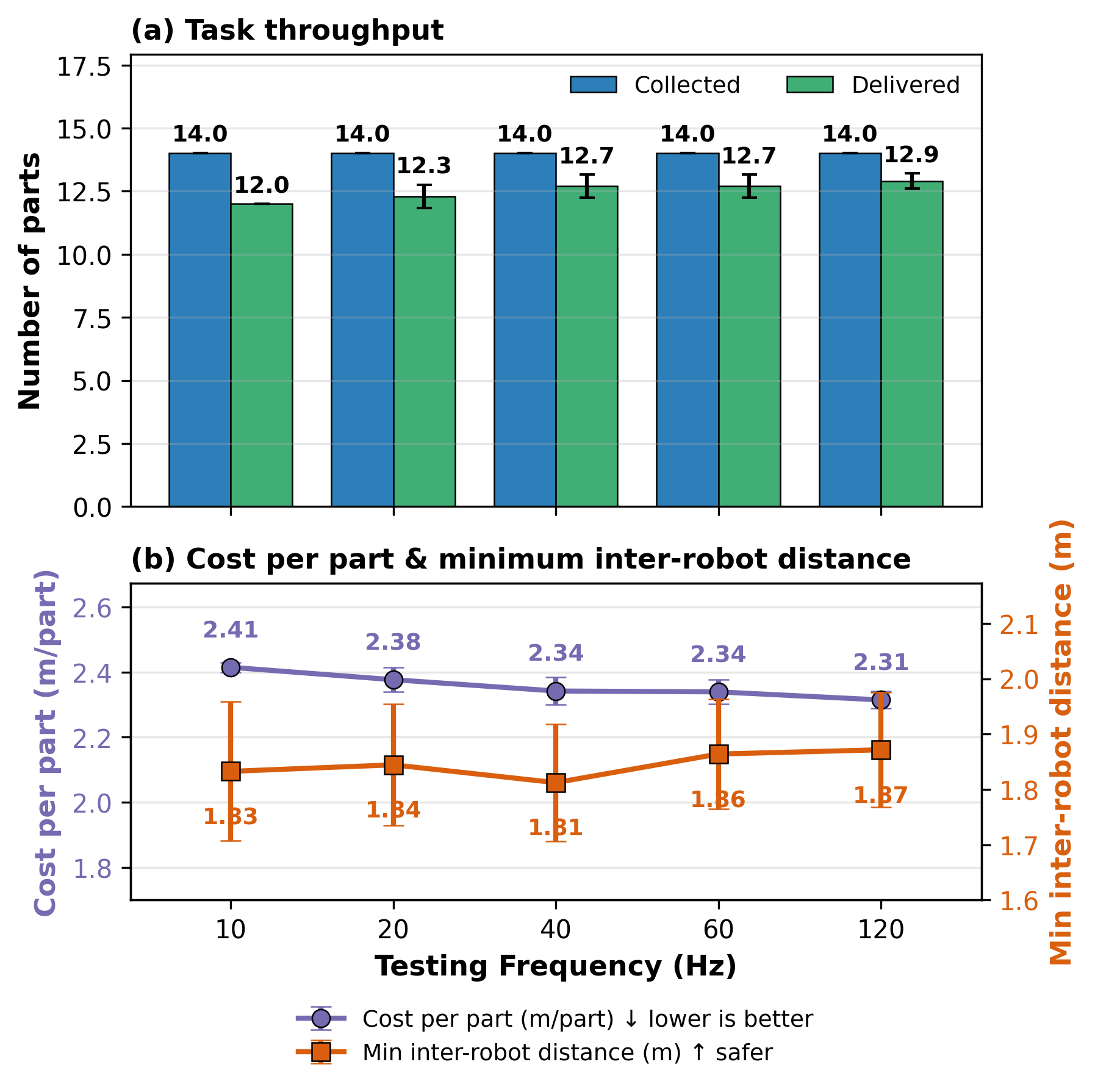}
    \includegraphics[width=0.485\linewidth]{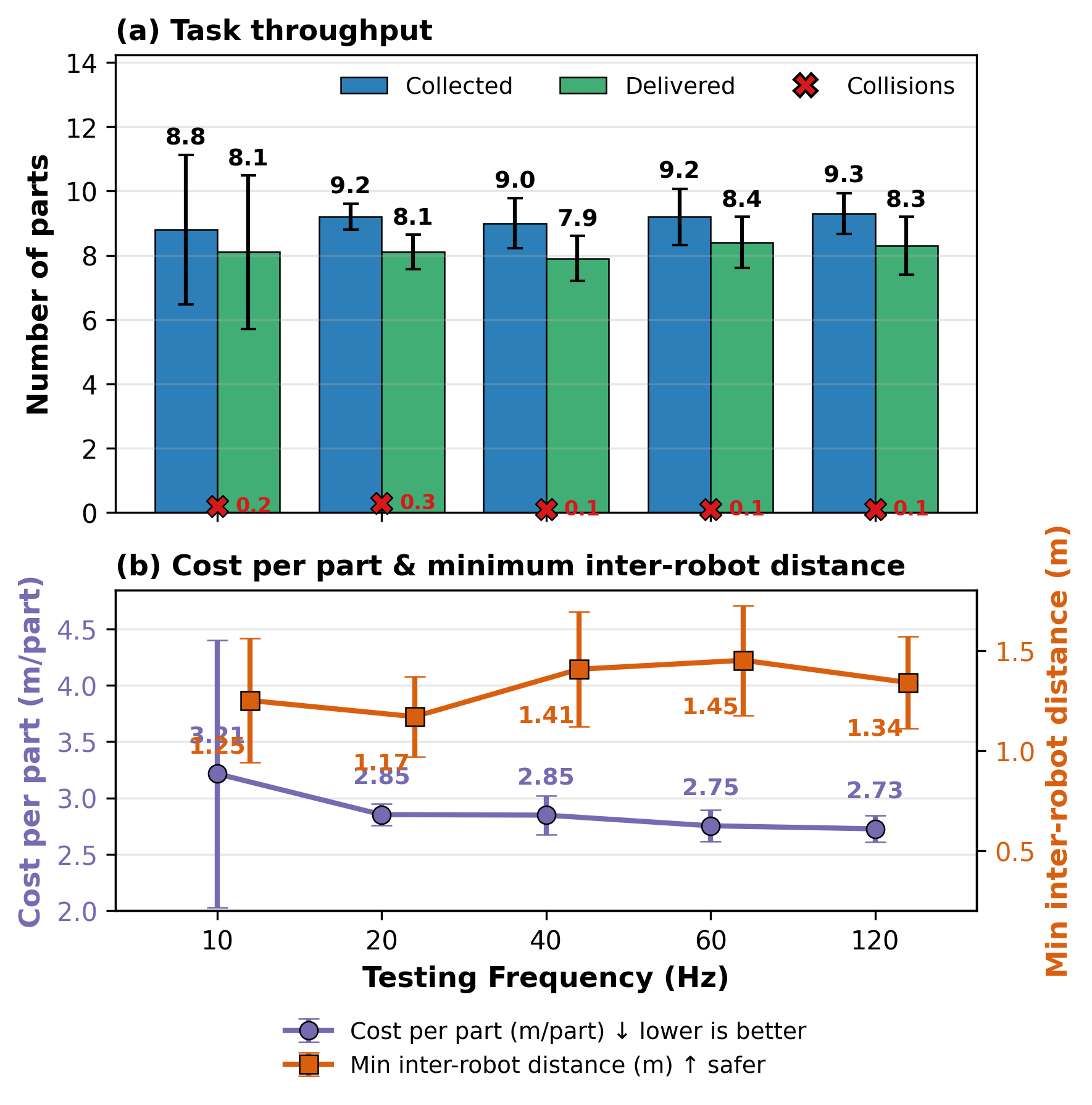}

    \caption{Position ($\sigma$ 20 cm), velocity ($\sigma$ 20 cm/s), LiDAR ($\sigma$ 1 cm), action ($\sigma$ 1 cm/s), and initial positions ($\sigma$ 20 cm), in the left Scenario 1, and in the right Scenario 2}
    \label{fig:freqP2_O2L01A01}
\end{figure*}

\begin{figure*}[!t]
    \centering
    \includegraphics[width=0.499\linewidth]{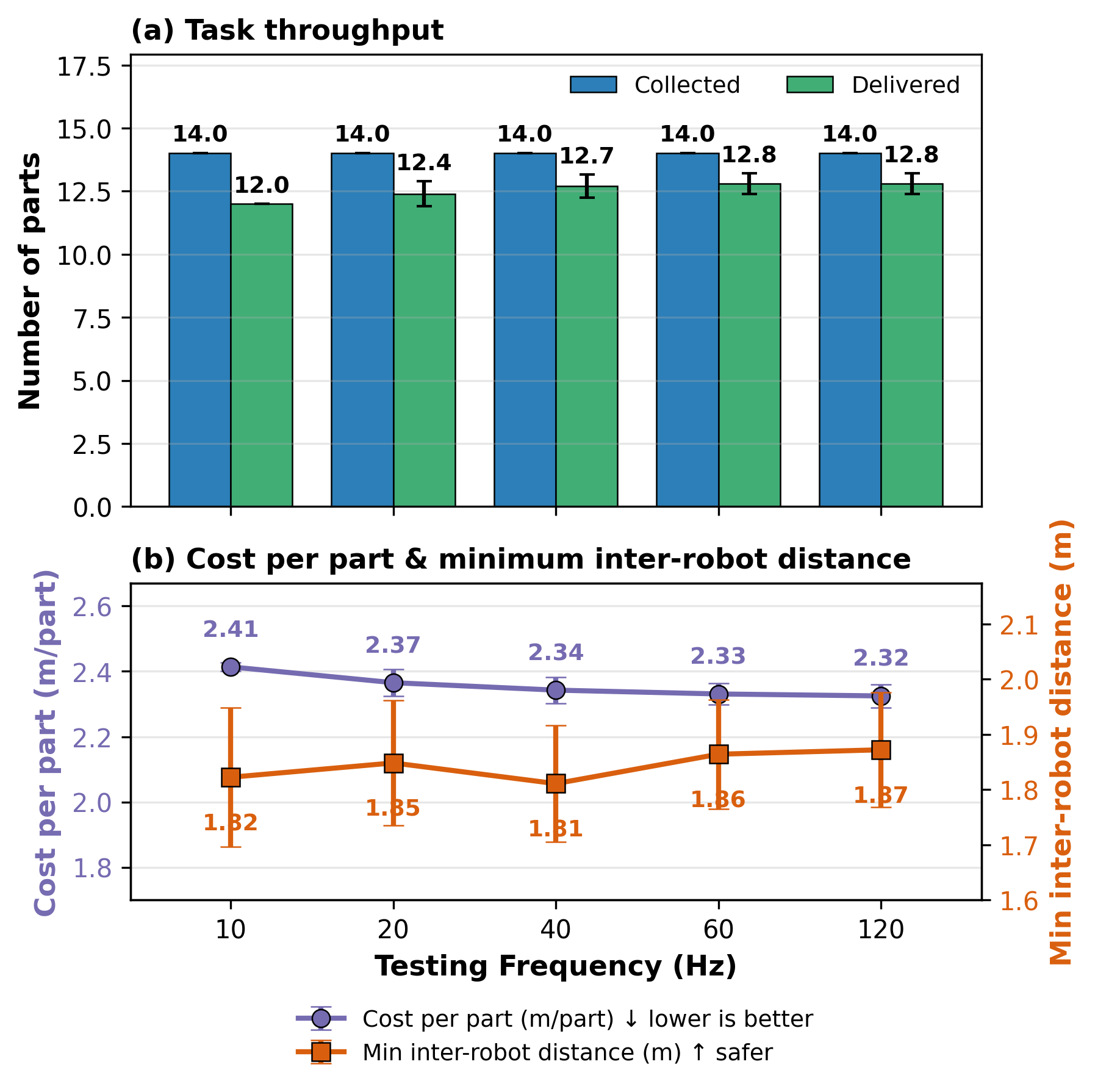}
    \includegraphics[width=0.485\linewidth]{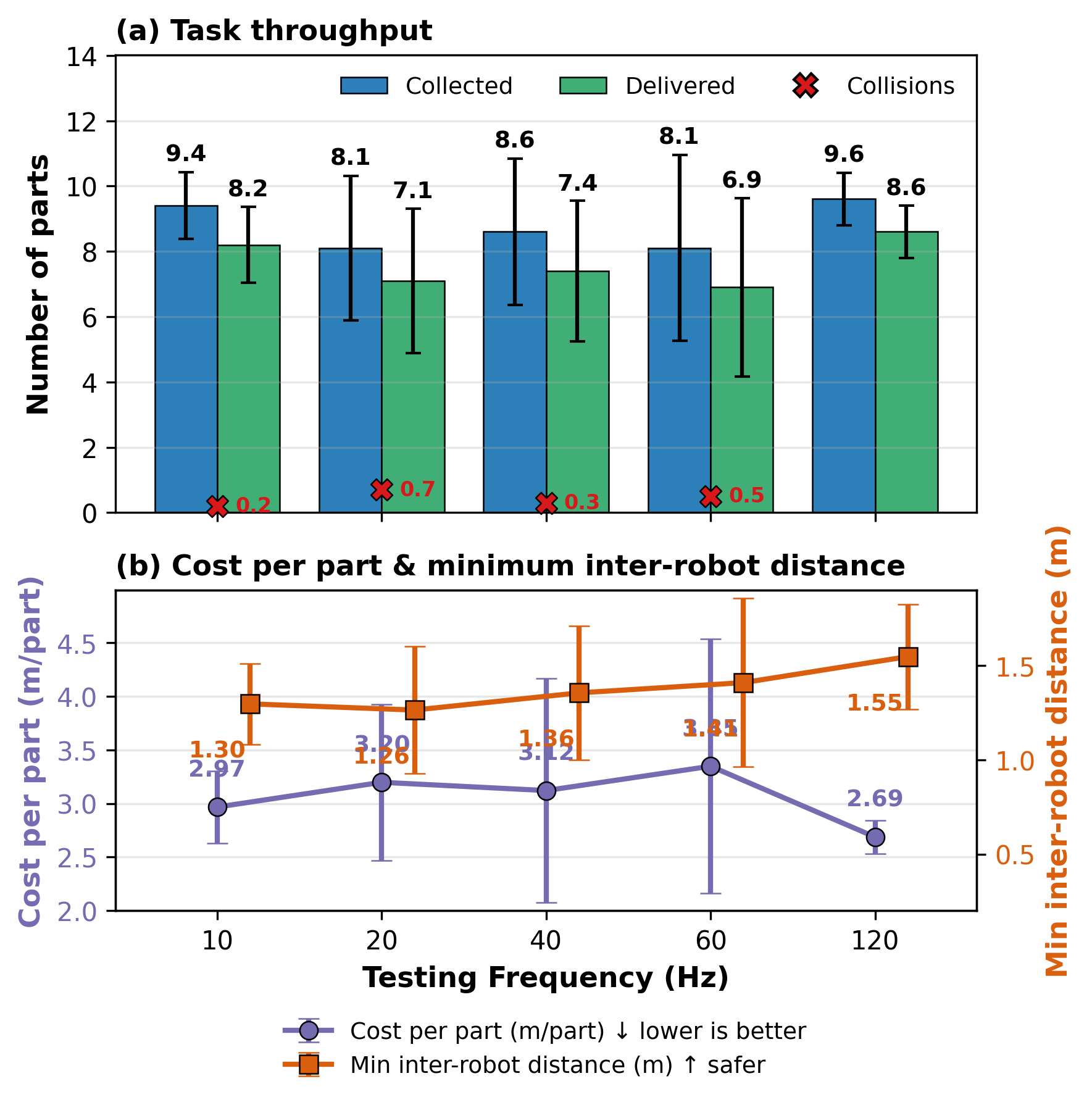}

    \caption{Position ($\sigma$ 10 cm), velocity ($\sigma$ 20 cm/s), LiDAR ($\sigma$ 1 cm), action ($\sigma$ 1 cm/s), and initial positions ($\sigma$ 20 cm), in the left Scenario 1 and in the right Scenario 2}
    \label{fig:P2_O1V2L01A01}
\end{figure*}

Here we provide the results of more experiments to investigate the effect of command update frequency on the performance under different noises level. It serves as a complementary experiment to the one presented in the paper, but also shows the robustness of FMAPPO under varying conditions. Zero-mean Gaussian noise was added to position and velocity estimations, LiDAR readings, action commands, and the robots' initial positions. Figures \ref{fig:freqP1_O02L01A01},\ref{fig:freqP2_O2L01A01} \ref{fig:P2_O1V2L01A01} report the results at different noise levels.

\end{document}